\documentclass{article}

\usepackage{microtype}
\usepackage{graphicx}
\usepackage{booktabs} 
\usepackage[utf8]{inputenc} 
\usepackage[T1]{fontenc} 
\usepackage{amsmath}
\usepackage{hyperref}
\usepackage{url} 
\usepackage{amsfonts} 
\usepackage{nicefrac}
\usepackage{microtype}
\usepackage{lipsum}
\usepackage{subcaption}
\usepackage[sort]{natbib}
\usepackage{tikz-cd}
\usepackage{enumitem, kantlipsum}
\graphicspath{ {./images/} }

\usepackage{hyperref}

\usepackage[accepted]{icml2021}

\DeclareMathOperator*{\argmax}{arg\,max}
\DeclareMathOperator*{\argmin}{arg\,min}

\newcommand{\norm}[1]{\left\lVert#1\right\rVert}

\begin{document}

\twocolumn[
\icmltitle{Adversarial Perturbations Are Not So Weird: Entanglement of Robust and Non-Robust Features in Neural Network Classifiers}



\icmlsetsymbol{equal}{*}

\begin{icmlauthorlist}
\icmlauthor{Jacob M. Springer}{lanl}
\icmlauthor{Melanie Mitchell}{sfi}
\icmlauthor{Garrett T. Kenyon}{lanl}
\end{icmlauthorlist}

\icmlaffiliation{lanl}{Los Alamos National Laboratory, Los Alamos, NM}

\icmlaffiliation{sfi}{Santa Fe Institute, Santa Fe, NM}

\icmlcorrespondingauthor{Jacob M. Springer}{jacmspringer@gmail.com}

\icmlkeywords{Machine Learning, Robustness, Non-robust, Features, Deep Learning, Neural Networks, Adversarial Examples, ICML}

\vskip 0.3in
]



\printAffiliationsAndNotice{}  

\begin{abstract}
Neural networks trained on visual data are well-known to be vulnerable to often imperceptible adversarial perturbations.  The reasons for this vulnerability are still being debated in the literature. Recently \citet{ilyas2019adversarial} showed that this vulnerability arises, in part, because neural network classifiers rely on highly predictive but brittle ``non-robust'' features.  In this paper we extend the work of Ilyas et al.\ by investigating the nature of the input patterns that give rise to these features.  In particular, we hypothesize that in a neural network trained in a standard way, non-robust features respond to small, ``non-semantic'' patterns that are typically entangled with larger, robust patterns, known to be more human-interpretable, as opposed to solely responding to statistical artifacts in a dataset.  Thus, adversarial examples can be formed via minimal perturbations to these small, entangled patterns. In addition, we demonstrate a corollary of our hypothesis: robust classifiers are more effective than standard (non-robust) ones as a source for generating transferable adversarial examples in both the untargeted and targeted settings.  The results we present in this paper provide  new insight into the nature of the non-robust features responsible for adversarial vulnerability of neural network classifiers.
\end{abstract}

\section{Introduction}

\begin{figure*}[ht]
    \captionsetup[subfigure]{labelformat=empty}
    \centering
    \begin{subfigure}[t]{0.28\textwidth}
    \centering
    \includegraphics[width=0.73\textwidth]{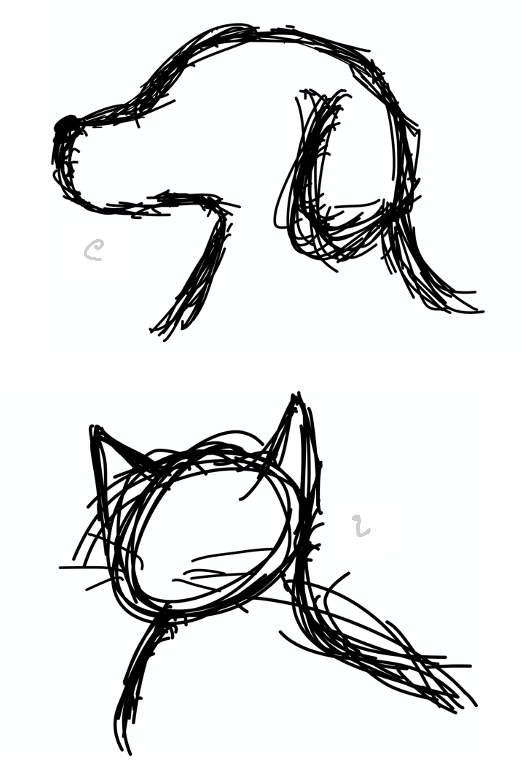}
    \caption{\textbf{Type A:} Robust features. These features commonly respond to human-interpretable (``semantic'') patterns (here of a dog or a cat). }
    \label{subfig:typea}
    \end{subfigure}
    \hspace{0.5em}
    \begin{subfigure}[t]{0.34\textwidth}
    \centering
    \includegraphics[width=1.0\textwidth]{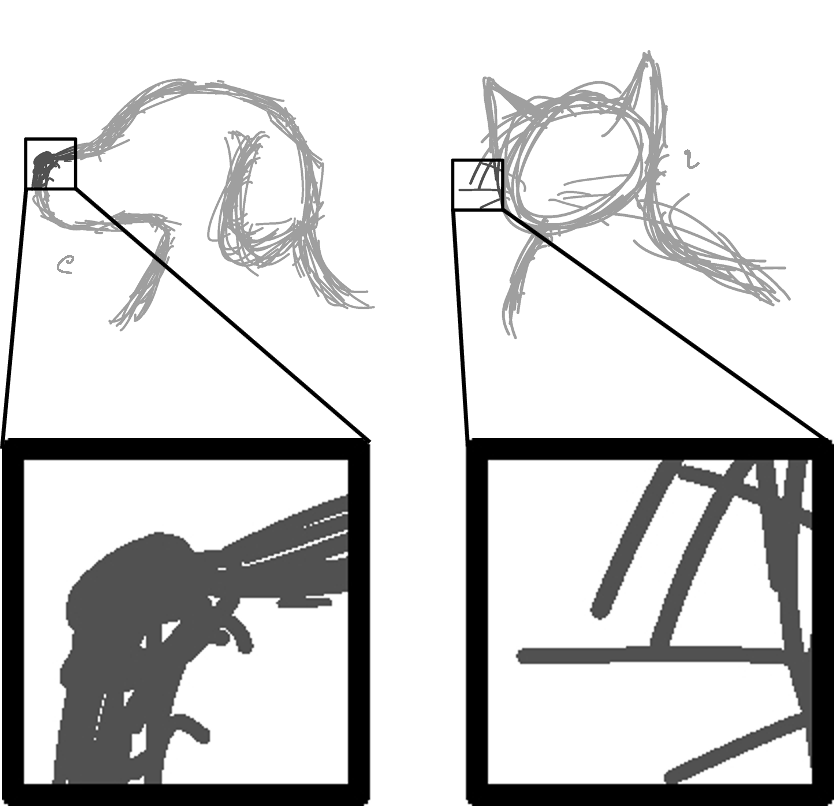}
    \caption{\textbf{Type B:} Non-robust features that respond to small yet highly predictive patterns that by themselves appear non-semantic, yet are entangled with patterns associated with robust features (e.g., in (a)).}
    \label{subfig:typeb}
    \end{subfigure}
    \hspace{0.5em}
    \begin{subfigure}[t]{0.34\textwidth}
    \centering
    \includegraphics[width=1.0\textwidth]{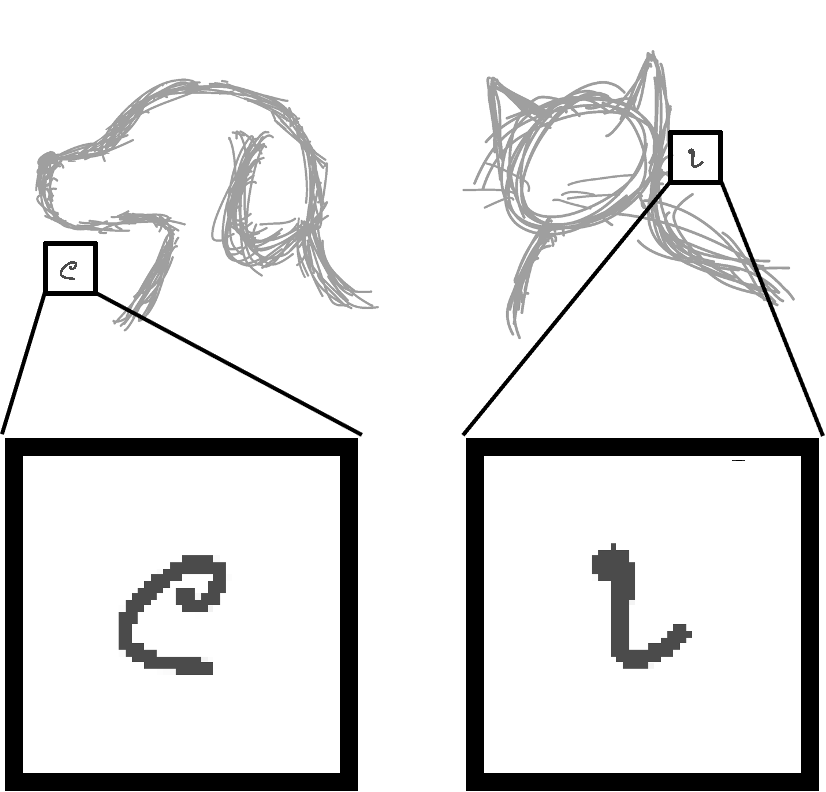}
    \caption{\textbf{Type C:} Non-robust features that respond to highly predictive patterns that are artifacts in the dataset, and are independent of robust features.}
    \label{subfig:typec}
    \end{subfigure}
    \caption{A sketch of the possible relationships between robust and non-robust features. Note that this is oversimplified as non-robust features could respond to combinations of (b) and (c).}
    \label{fig:types_of_features}
\end{figure*}

It is well-known that neural network classifiers trained on visual data are susceptible to adversarial examples---images that have been minimally perturbed so as to look unchanged to humans but are classified incorrectly, even though the original image is correctly classified.  Many explanations have been offered for this susceptibility as well as for the transferability of adversarial examples across network architectures and even training sets; however, the ML community's understanding of these phenomena remains incomplete.

In this paper we extend the work of \citet{ilyas2019adversarial} on understanding these phenomena as the result of highly predictive but brittle \textit{non-robust} features that are learned by neural networks undergoing standard supervised training.  Ilyas et al.\ proposed that adversarial examples are not ``bugs'' resulting from properties of a network that cause odd behavior on examples that are off the training manifold, but rather that they are due to learned ``features'' that respond to predictive---yet non-interpretable---patterns in the dataset.  Ilyas et al.\ showed that such non-robust features correspond to patterns that are widespread in the data.  Moreover, they showed that these non-robust features can account for much if not all of the high accuracy of neural networks on image datasets such as CIFAR-10 and ImageNet-9, and that perturbations targeting these non-robust features allow for transferable adversarial attacks.

Here, we investigate the nature of these non-robust features.  In particular, we give empirical evidence that in a neural network trained in a standard way, predictive but non-robust features respond to small patterns that are typically entangled with larger, human-interpretable patterns, as opposed to responding to separate statistical artifacts in a dataset.  We argue that the non-robust features of neural networks that can be exploited to construct seemingly uninterpretable adversarial perturbations in fact rely on the same underlying patterns captured by more robust features. While we use many of the experimental methods proposed by \citet{ilyas2019adversarial} as tools, the question which we seek to answer and our results are distinct: we identify a relationship between robust and non-robust features and then demonstrate that our finding can motivate highly effective targeted transferable adversarial examples. Ilyas et al., on the other hand, consider robust and non-robust features separately.

By demonstrating that a classifier for which only non-robust features are useful can achieve well-above-chance accuracy on a test set in which only robust features are useful, we confirm that networks learn non-robust features that are closely entangled with robust features. Robust features have been shown to be substantially more interpretable than non-robust features \citep{engstrom2019adversarial, santurkar2019image, kaur2019perceptually}. By establishing the relationship between non-robust and robust features, we argue that non-robust features may be more interpretable than they appear through the lens of adversarial perturbations. In other words, non-robust features \textit{may not be so weird, after all.}

The results of this paper have important implications. We explain conceptually why adversarial perturbations might appear non-semantic yet are related to semantic features. We present and test a corollary of our results: transferable examples (targeted and untargeted) can be generated more effectively by using robust classifiers rather than standard classifiers. Finally, our results provide new insights into the nature of adversarial vulnerabilities, and suggest directions of future research.

\section{Terminology}

In this section, following \citet{ilyas2019adversarial}, we define several important terms, in particular the notions of \textit{robust} and \textit{non-robust} features.

Deep learning classifiers, in particular convolutional neural networks (CNNs) are typically composed of a sequence of non-linear transformations called layers, the result of which is fed through a final linear classifier layer to select a class \citep{lecun1998gradient, krizhevsky2012imagenet}. We refer to the penultimate layer as the \textit{representation layer}, denoted $\mathit{rep}(x)$, where $x$ is the $n$ dimensional input to the network (e.g., an image).  We refer to each unit of the representation layer as a \textit{feature} that the network computes for classification. Each feature $f: \mathbb{R}^n \to \mathbb{R}$ maps the input to a real number---the feature's activation, given the input.

We adapt the following definitions from \citet{ilyas2019adversarial}, who considered the binary classification case, in which the dataset $\mathcal{D}$ consists of pairs $(x,y) \in \mathcal{D}, x \in \mathbb{R}^n, y \in \{1,-1\}$, and the final layer outputs either $1$ or $-1$.

\begin{itemize}[leftmargin=*, nosep]
	
\item A feature $f$ is \textit{useful} for a dataset $\mathcal{D}$ when there exists a $\rho > 0$ such that, $$\mathbb{E}_{(x, y) \in \mathcal{D}}[y \cdot f(x)] \geq \rho,$$ or, more intuitively, when $f$ is correlated with the class $y$ of an input $x$.
	
\item For a given input $x$, adversarial example $\hat{x} = x + \delta$, and $\varepsilon > 0$, $\delta$ is said to be a \textit{permissible perturbation} when $\norm{\delta}_2 < \varepsilon$.
  
\item A feature $f$ is \textit{robust} when, for some $\gamma > 0$,

\[\mathbb{E}_{(x, y) \in \mathcal{D}}[\min_{\norm{\delta}_2 \leq \varepsilon} y \cdot f(x + \delta)] \geq \gamma,\]

More intuitively, a feature is \textit{robust} when it is correlated with the class $y$ even under worst-case permissible perturbations of the input $x$.
	
\item For simplicity, a feature is said to be \textit{non-robust} when it is useful but not \textit{robust}. (Here we do not consider non-useful features.)
	
\end{itemize}

A feature is a property of a classifier, and describes a way in which the classifier measures information in the input. However, it will also be conceptually useful to refer to the information itself. Thus we define the closely related notion of a \textit{pattern} $P \subset \mathbb{R}^n$, a subset of inputs. We say that an image $x$ \textit{contains} a pattern $P$ when $x \in P$. For example, a pattern of stop signs would be the set of all images that contain a stop sign. 

We are primarily concerned with the relationship between robust features and non-robust features.

\subsection{Hypotheses Regarding The Relationship Between Robust and Non-Robust Features}

The main question we address in this paper is, what is the nature of the \textit{non-robust} features described by \citet{ilyas2019adversarial}?  We hypothesize that many \textit{non-robust} features learned by standard (i.e., non-robust classifiers) respond to patterns that are entangled with human-interpretable \textit{robust} features, rather than responding to dataset artifacts.  Moreover, these entangled non-robust features can be exploited to create adversarial examples.  If true, this hypothesis has important implications for defenses against, and transferability of, adversarial examples.

We illustrate our hypothesis with a conceptual diagram in Figure~\ref{fig:types_of_features} consisting of three different types of patterns that features in a network could respond to.  Robust features respond to Type A patterns---ones that are interpretable by humans as giving rise to a particular class.  Type C illustrates a type of pattern that non-robust features might respond to: highly predictive non-semantic patterns in the dataset that are unrelated to the robust features in the image.  These might be called ``spurious correlations'' or ``artifacts'' in the dataset.  Type B illustrates another possibility for non-robust features: they might respond to small but highly predictive components of robust features (e.g., the nose of a dog or the whiskers of a cat) that could be easily exploited to yield adversarial examples.

We confirm in this paper that Type~B features are indeed learned by CNNs, that these features can, in some cases, explain the high accuracy of CNNs, and that these features can be exploited in adversarial attacks.  Thus the adversarial vulnerability of CNNs is not necessarily due to non-semantic artifacts, but can be explained in terms of non-robust features that are entangled with robust features that respond to semantically meaningful patterns.

\section{Methods \label{sec:methods}}

\subsection{Strategy for Testing Our Hypotheses}

In order to test our hypothesis that Type B features are learned by standard CNN classifiers, we use the following strategy, inspired by \citet{ilyas2019adversarial}.  We construct a neural network classifier in which only non-robust features are useful.  We then construct a new test set for which only robust features are useful for classification.

We then show that that the classifier achieves substantially higher than chance accuracy on the constructed test set; this implies that the classifier must be using non-robust features that are entangled with robust features---namely, Type B features.

The following subsections describe how we construct this classifier and test set. 

\subsection{Constructing a Classifier For Which Only Non-Robust Features Are Useful \label{sec:classifierconstruction}}

\begin{figure}
    \centering
    \includegraphics[width=\linewidth, trim = 0cm 1.4cm 0cm 2.15cm, clip]{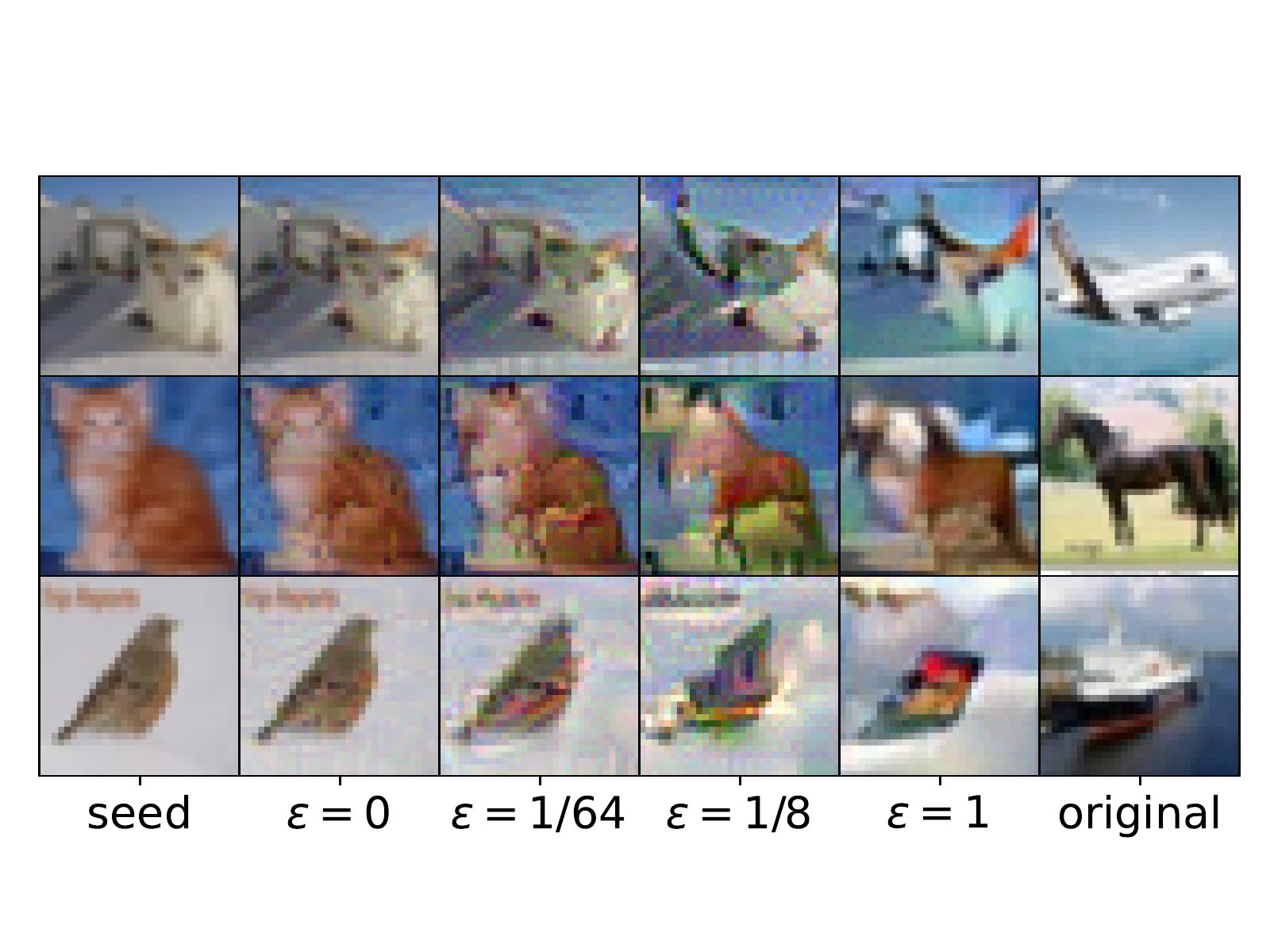}
    \caption{Examples of robustified images by $\varepsilon$ of robust classifier. The rightmost column depicts the original CIFAR-10 image; the leftmost column depicts the seed ($x_0$) from which the gradient descent of the robustification process began; the rest are robustified images. Since the representation layers of less robust classifiers are more easily perturbed, robustified images generated with a less robust model (smaller value of $\varepsilon$) will appear closer to the (random) CIFAR-10 image that seeded the gradient-descent process. (Best viewed in color.)}
    \label{fig:robustification_examples}
\end{figure}

\paragraph{Zero-Robust Classifier}


We use the term \textit{zero-robust classifier} for the classifier for which only non-robust features are useful. To train these classifiers, we follow the method of \citet{ilyas2019adversarial}. We first construct a training set $\mathcal{D}_{\text{zero}}$ from our original training set $\mathcal{D}$ such that under $\mathcal{D}_{\text{zero}}$, all robust features are non-useful, i.e., totally uncorrelated with the assigned label, while non-robust features remain useful. By training a standard classifier on this distribution, the standard classifier should only learn non-robust features.

The new training set $\mathcal{D}_{\text{zero}}$ can be constructed as follows: for each input-label pair $(x, y) \in \mathcal{D}$, choose a new label $\hat{y}$ uniformly at random from the possible labels. Construct a permissible adversarial example $\hat{x}$ (i.e., $\hat{x} = x + \delta$, where $\delta$ is a permissible perturbation) such that $\hat{x}$ is classified as $\hat{y}$. Then, let $(\hat{x}, \hat{y})$ be an element of $\mathcal{D}_{\text{zero}}$. \citet{ilyas2019adversarial} argued that there should be almost no correlation between robust features and the assigned labels in this new dataset. Thus, a classifier trained on this dataset should not rely on robust features.

\paragraph{Negative-Robust Classifier:  Addressing Problem of ``Robust Feature Leakage''.}
As pointed out by \citet{engstrom2019a}, there may be a small correlation between robust features and the assigned labels in $\mathcal{D}_{\text{zero}}$, due to ``robust feature leakage,'' which may account for some of the accuracy of a zero-robust classifier on a robustified dataset. We control for this possibility with the method proposed by \citet{ilyas2019adversarial}: we construct a \textit{negative-robust} classifier, where robust features are anti-correlated with the label and non-robust features are correlated with the label. Thus, robust features should actively hurt accuracy and any positive accuracy can certainly be attributed to non-robust features. We construct a negative-robust classifier similar to the construction of the zero-robust classifier, but while $\hat{y}$ is selected uniformly at random to construct $\mathcal{D}_{\text{zero}}$, for the negative-robust classifier we deterministically permute the class labels, associating each label $y$ with the corresponding $\hat{y}$ in the permutation.  Then for each $(x,y)$ pair in $\mathcal{D}$ we create $(\hat{x},\hat{y})$, where $\hat{x}$ is an adversarial example based on $x$ targeting class $\hat{y}$.  Using a deterministic permutation of the label classes makes robust features {\it anti-correlated} with the correct label class.

\subsection{Constructing a Test Set For Which Only Robust Features Are Useful \label{sec:robustification}}

In order to construct a test set for which only robust features are useful, we follow the approach of \citet{ilyas2019adversarial}.  First we construct a \textit{robust classifier}, and then use that classifier to construct the desired dataset. 

\paragraph{Robust Classifier.}
We construct robust classifiers via standard adversarial training \cite{madry2017towards}. Formally, we wish to find classifier parameters $\theta^{*}$ that minimize the following expression:

\[ \theta^{*} = \argmin_\theta \mathbb{E}_{(x, y) \in \mathcal{D}}[ \max_{\norm{\delta}_2 \leq \varepsilon}\mathcal{L}(\theta, x + \delta, y)] \]

where $\mathcal{L}(\theta, x, y)$ computes the cross-entropy loss of the classifier with parameters $\theta$ on the input $x$ and label $y$. The magnitude of adversarial examples to which the classifier is robust is characterized by the $\varepsilon$ parameter, which we call the \textit{robustness parameter} of a classifier.  Larger values of $\varepsilon$ correspond to more robust classifiers (i.e., they are robust to larger perturbations).  The special case when $\varepsilon = 0$ corresponds to a standard classifier.  We refer to a classifier with robustness parameter $\varepsilon$ as an $\varepsilon$\textit{-robust classifier}. 

The above saddle-point optimization problem can be solved by first solving the inner maximization problem with projected gradient ascent and then solving the outer maximization problem with standard back-propagation. We train with projected gradient descent as proposed by \citet{madry2017towards}. Since this problem is well studied, we defer to prior literature for discussion of the matter \citep{madry2017towards, carlini2019evaluating}.

\paragraph{Test Set Robustification.}
To construct a test set for which only robust features are useful for classification, we train a robust classifier $C$ as described above, and use it to create a new test set $\hat{\mathcal{D}}$ in which there is a one-to-one mapping between the elements of $\hat{\mathcal{D}}$ and the original test set $\mathcal{D}$: $(x,y) \mapsto (\hat{x}, y)$. In particular, an image $\hat{x} \in \hat{\mathcal{D}}$ is constructed so that any feature of $C$ that is useful in classifying $x$ is equally useful in classifying $\hat{x}$, in the sense that useful features are equally correlated with the label $y$, and no other possible feature is useful.  In short, our construction will ensure that the \textit{only} useful features for classifying $\hat{x}$ will be the features of C, which are, by definition, robust features.

Define $\mathit{rep_C}(x)$ as the output of the $m$-dimensional representation (penultimate) layer of robust classifier $C$ with input $x$. Then for $r \in \mathbb{R}^m$, $\mathit{rep_C}^{-1}(r)$ is a set of images with robust features $r$.

We cannot easily compute $\mathit{rep_C}^{-1}(r)$. However, for each element of $x$ our original test set, we can find an approximate inverse by solving the minimization problem 
\[ \min_{x_r} \norm{\mathit{rep_C}(x_r) - \mathit{rep_C}(x)}. \]  We solve this via gradient descent, starting from an image $x_0$. In other words, starting from randomly chosen test-set image $x_0$, we search for an image $x_r$ whose representation $\mathit{rep_C}(x_r)$ is as close as possible to $\mathit{rep_C}(x)$.

Following \citet{ilyas2019adversarial}, if we choose the initial image $x_0$ uniformly from the test set, and the test set has a uniform distribution over labels, then all features in $\mathit{rep_C}(x_0)$ are uncorrelated (in expectation over $x_0$) with $x$'s label.  This ensures (in expectation) that any features that respond to patterns in $x_0$ will not be correlated with $x$'s label, $y$.  Since we apply gradient descent to $x_0$ in order to find an $\hat{x}$ whose representation is similar as possible to $C$'s representation for $x$, the only features that correlate with $y$ will be those in $C$, which are robust by definition.  Thus only robust features are useful for classifying $\hat{x}$.

Following \citet{ilyas2019adversarial}, we refer to the inversion process the \textit{robustification} of images.  Given a test set $\mathcal{D}$, we can construct a new test set $\hat{\mathcal{D}}$ by computing this approximation of $\mathit{rep}^{-1}(\mathit{rep}(x))$ and labeling it $y$ for every $(x, y) \in D$.

If robust classifier $C$ has robustness parameter $\varepsilon > 0$, we say that $\hat{\mathcal{D}}$ is the $\varepsilon$-\textit{robustification} of $\mathcal{D}$ and will be referred to as $\mathcal{R}_\varepsilon$.  Figure~\ref{fig:robustification_examples} shows examples of the robustification process on CIFAR-10 test images. 

\subsection{Datasets \& Model Consideration}

We evaluate our methods on two datasets: CIFAR-10 \citep{krizhevsky2009learning} and ImageNet-9, a derivative of ImageNet \citep{ILSVRC15} with two key differences from the original ImageNet dataset: first, we use a $64 \times 64$ pixel downsampled version of the ImageNet data for computational feasibility \citep{chrabaszcz2017downsampled}, second, adopted from \citet{ilyas2019adversarial}, we reduce the number of classes to nine: dog, cat, frog, turtle, bird, primate, fish, crab, insect. See \citet{ilyas2019adversarial} for further details.

For the CIFAR-10 dataset, we train ResNet50 classifiers \citep{he2016deep, he2016identity}. However, due to the high computational cost of training robust models on ImageNet-9, we train a classifier with fewer parameters, ResNet20, on the ImageNet-9 dataset. 

\section{Results of Experiments}

\begin{table}[h]
\centering
\caption{Accuracy of zero-robust, negative-robust, standard, and robust ResNet50 classifiers on robustified CIFAR-10 test set. All classifiers are trained starting with random initial weights. The \textit{Test} column is the accuracy on the original CIFAR-10 test set. The $\mathcal{R}_\varepsilon$ columns give accuracy on robustified CIFAR-10 test set, generated with respect to a robust ResNet50 with robustness parameter $\varepsilon$.}
\label{tab:performance_robustified}
\begin{tabular}{@{}lllllll@{}}
\toprule
Classifier & Test & $R_0$ & $\mathcal{R}_{\nicefrac{1}{16}}$ & $\mathcal{R}_{\nicefrac{1}{4}}$ & $\mathcal{R}_{\nicefrac{1}{2}}$ & $\mathcal{R}_{1}$ \\ \midrule
Standard & 94.1 & 58.3 & 79.9 & 75.1 & 69.2 & 56.3 \\
Robust & 86.5 & 9.93 & 18.2 & 70.5 & 86.5 & 70.0 \\
Zero-robust & 46.8 & 38.8 & 23.3 & 22.1 & 23.4 & 21.4 \\
Neg-robust & 21.9 & 31.8 & 11.1 & 12.8 & 14.3 & 13.2 \\
\bottomrule
\end{tabular}
\end{table}

\begin{table}[h]
\centering
\caption{Accuracy of zero-robust, negative-robust, standard, and robust ResNet20 classifiers on robustified ImageNet-9 test set. Analogous to Table~\ref{tab:performance_robustified}. Note: since there are nine classes, chance accuracy is approximately $11.1\%$.}
\label{tab:performance_robustified_imagenet}
\begin{tabular}{@{}lllllll@{}}
\toprule
Classifier & Test & $R_0$ & $\mathcal{R}_{\nicefrac{1}{16}}$ & $\mathcal{R}_{\nicefrac{1}{4}}$ & $\mathcal{R}_{\nicefrac{1}{2}}$ & $\mathcal{R}_{1}$ \\ \midrule
Standard & 94.8 & 51.9 & 76.3 & 63.4 & 54.6 & 47.2 \\
Robust & 82.9 & 44.1 & 46.1 & 49.0 & 51.7 & 56.0 \\
Zero-robust & 49.2 & 24.8 & 24.3 & 24.6 & 28.6 & 31.4 \\
Neg-robust  & 38.3 & 25.4 & 23.2 & 21.8 & 24.4 & 23.7 \\
\bottomrule
\end{tabular}
\end{table}

Recall our definition of Type B features: Non-robust features learned by a network that respond to small yet highly predictive patterns that by themselves appear non-semantic, yet are entangled with patterns associated with robust features.  Our hypothesis is that Type B features are prevalent in standard (i.e., non-robust) image classifiers, and can be exploited to create adversarial examples. That is, Type C features (which respond to dataset artifacts) are not the only features responsible for adversarial vulnerability.

We present here the result of an experiment in which we evaluate zero-robust and negative-robust classifiers, representing classifiers that rely exclusively on non-robust features, on CIFAR-10 and ImageNet-9 test sets that have been robustified---that is, in which only robust features are useful.

The results, given in Tables~\ref{tab:performance_robustified}~and~\ref{tab:performance_robustified_imagenet}, confirm our hypothesis: the zero-robust classifier performs above chance on all robustified test sets; this can only occur when Type B features are present. Similarly, the negative-robust classifier, which acts as a control for the problem of robust feature leakage (see Section~\ref{sec:classifierconstruction}), has above-chance accuracy on robustified test sets except for when $\varepsilon$ is large. We expect this result, since the presence of robust features decreases accuracy in negative-robust classifiers. In fact, in the absence of Type B features, we would expect the negative-robust classifier to perform at below-chance accuracy on the robustified test sets. As above-chance accuracy by the negative-robust can only be explained by the presence of Type B features, the accuracy of the negative-robust classifier serves as a lower-bound on the accuracy for which can be accounted by Type B features. The fact that the accuracy is above chance for many of the robustified test sets further confirms our hypothesis, and rejects the possibility that robust feature leakage entirely explains the above-chance accuracy of the zero-robust classifier on the robustified test sets. We include the accuracies of a standard classifier and an example robust classifier with robustness parameter $\varepsilon=0.5$ for comparison with the zero- and negative-robust classifier results.

\begin{figure}[t]
    \centering
    \includegraphics[width=\linewidth]{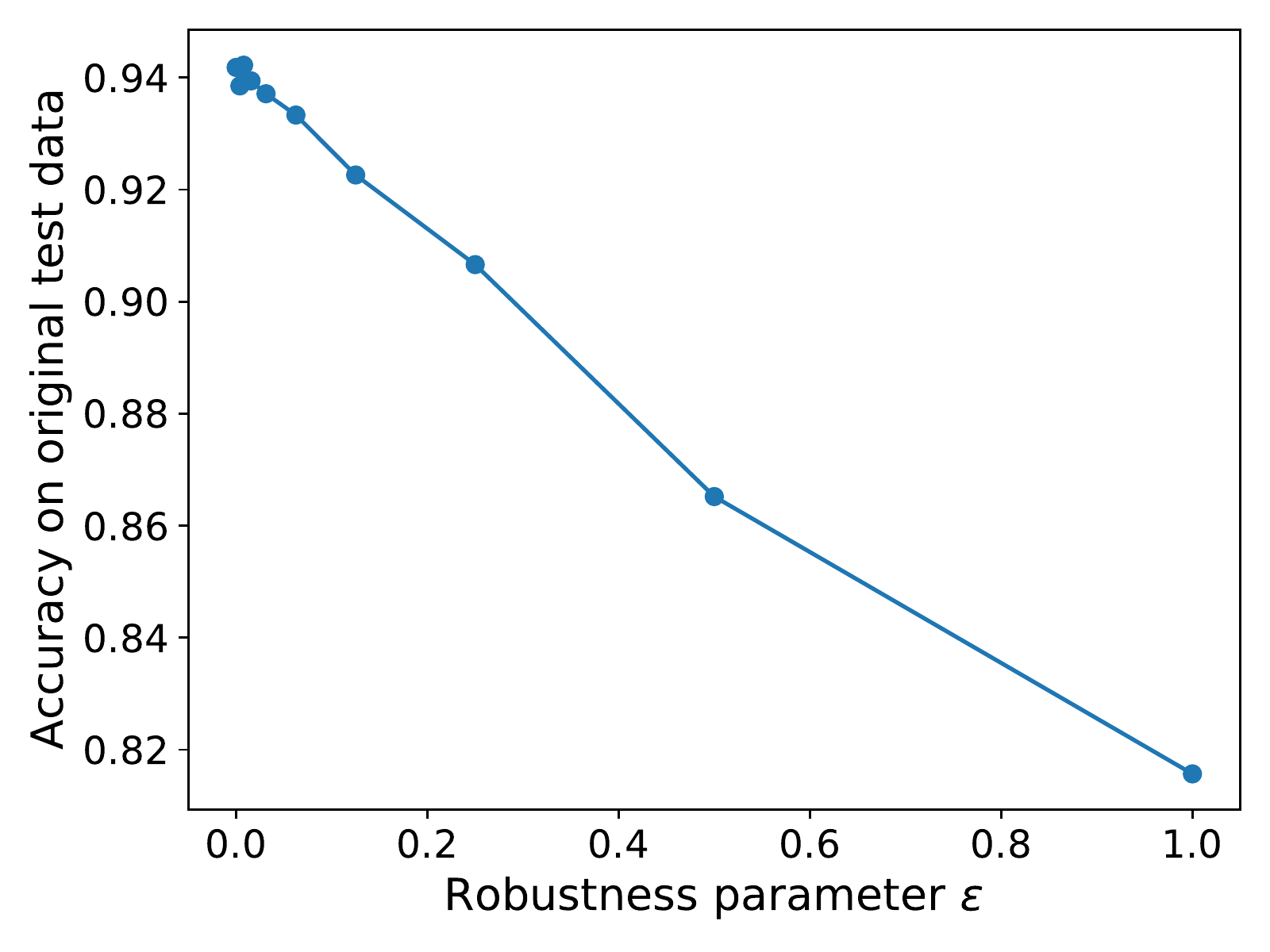}
    \caption{Accuracy of robust ResNet50 classifiers trained on the CIFAR-10 training data and evaluated on the CIFAR-10 original test data. See Appendix for the analogous graph for ImageNet-9. (Best viewed in color.)}
    \label{fig:robust_test_accuracy}
\end{figure}

We observe that the accuracy of zero-robust and negative-robust classifiers decreases on the robustified test sets as the robustness parameter $\varepsilon$ increases. We have already discussed that we expect this in the negative-robust classifier due to the negative influence of robust features on accuracy. However, we hypothesize that the reason zero-robust accuracy decreases is similar to that of the well-established result that the accuracy of robust classifiers decreases on the original test set as the robustness parameter of the classifier increases \citep{tsipras2019robustness, zhang2019theoretically}. We confirm this well-known result in Figure~\ref{fig:robust_test_accuracy}. We speculate that as the robustness increases, the classifier begins to ignore useful features, even some that are robust for smaller $\varepsilon$. Thus, the robustified test sets contain fewer useful features as $\varepsilon$ increases, and thus zero-robust accuracy decreases.

\begin{figure}[t]
\centering
\includegraphics[width=1.0\linewidth]{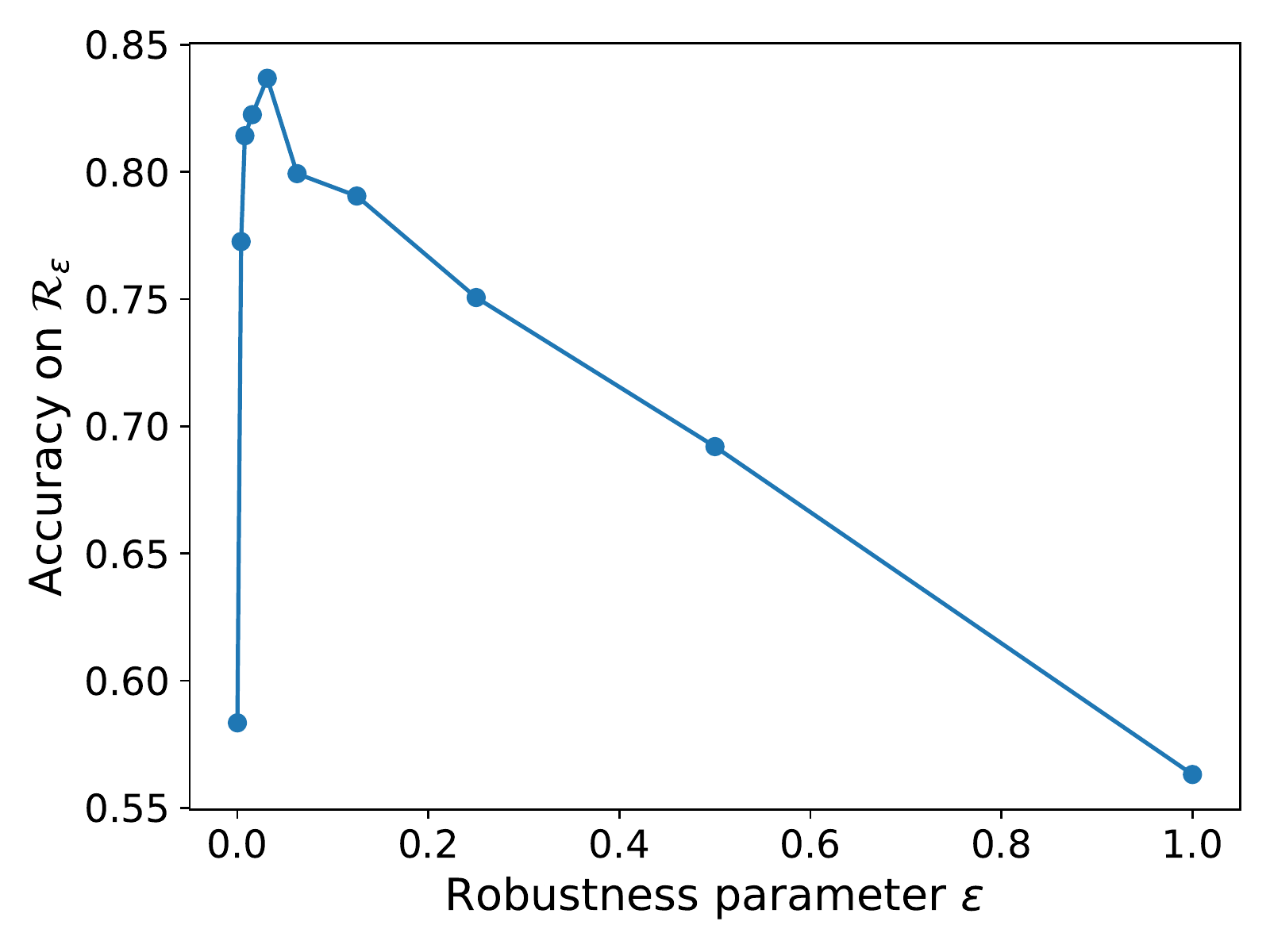}
\caption{Accuracy of a standard ResNet50 classifier on robustifications of the CIFAR-10 test dataset, generated with respect to robust ResNet50 classifiers with varying robustness parameter $\varepsilon$. See Appendix for the analogous graph for ImageNet-9. (Best viewed in color.)}
\label{fig:standard_performance_on_robustified}
\end{figure}

\section{The Universality of Robust Features}

The previous section established that non-robust features respond to patterns that are entangled with the patterns responded to by robust features.  However, there may be many equally predictive ``non-robust'' patterns that can be entangled with a single ``robust'' pattern.  For example, as illustrated in Figure~\ref{fig:types_of_features}, a non-robust Type~B feature which responds to the texture of a dog nose may be a component of a robust Type~A feature that identifies the shape of an entire dog. However, there may be many components of a particular robust feature that would equivalently predict ``dog''. Standard (non-robust) classifier $C_1$ might learn a particular subset of predictive non-robust (i.e., Type B) features, whereas another standard classifier $C_2$ with a different architecture or different initial weights might learn a different subset, with both classifiers exhibiting similar accuracy on a test set.  The features of both classifiers would be responding to \textit{different} subpatterns of the same underlying ``robust'' patterns; in this case, the features learned by $C_1$ would not strongly overlap the features learned by $C_2$.

The features learned by a classifier that tend to overlap with the features learned by other classifiers trained on the same dataset are characterized by \textit{universality}, as they are roughly the same across classifiers \citep{olah2020zoom}. \citet{li2015convergent} called this phenomenon ``convergent learning''.

We hypothesize that robust features will be more universal than non-robust features, since universal features likely represent the underlying, better generalizing properties of the dataset that are captured by robust features.  Analogous to the way we demonstrated the existence of Type~B features entangled with Type~A features in Section~\ref{sec:methods}, here we give evidence for the hypothesis that robust features are more universal than non-robust features by
evaluating standard classifiers on robustified test sets.

Recall that for a test set $D_T$ that has been robustified with respect to a robust classifier $C$, the only features that are useful for classification are robust features useful to $C$.  If a different classifier $C'$ has high accuracy on $D_T$, then $C'$ must be using the same (or very similar) features as $C$.  If many different classifiers have high accuracy on $D_T$, then we can say that $C$'s robust features are \textit{universal}.

\subsection{Comparing the Universality of Robust vs. Non-Robust Features}

Figure~\ref{fig:standard_performance_on_robustified} shows the accuracy of a standard (non-robust) ResNet50 classifier, trained on the CIFAR-10 training set, when tested on robustified CIFAR-10 test sets with varying robustness parameter $\varepsilon$.  The figure shows that the the standard classifier, which we'll call $C'$, has significantly higher accuracy on robustified test sets with $0 < \varepsilon \leq \nicefrac{1}{2}$, with a peak at $\varepsilon=\nicefrac{1}{32}$.  

To show why this supports the hypothesis that robust features are more universal than non-robust features, let $C_{\varepsilon}$ denote the $\varepsilon$-robust ResNet50 classifier used to create the robustified test set $R_{\varepsilon}$. Figure~\ref{fig:standard_performance_on_robustified} shows that the accuracy of $C'$ is low on $R_{0}$, which was designed specifically so that the features of $C_{0}$ would be useful to classify it.  This means that $C_{0}$'s features, which are likely to be mostly non-robust since $\varepsilon = 0$, tend not to be shared by $C'$.  However, the accuracy of $C'$ on $R_{\varepsilon}$ for $0 < \varepsilon \leq \nicefrac{1}{2}$ is dramatically higher, meaning that the features that are useful to classifiers $C_{\varepsilon}$ are also useful to $C'$, for $\varepsilon$ in that range.  Thus the robust features useful to these $C_{\varepsilon}$ tend to be shared by $C'$. Repeated evaluations with different initial weights for $C'$ exhibited the same behavior. Thus, the results shown in Figure~\ref{fig:standard_performance_on_robustified} support the hypothesis that robust features are more universal than non-robust features.

It is noteworthy that accuracy is highest for a relatively small robustness parameter of $\varepsilon = \nicefrac{1}{32}$.  We hypothesize that as robustness increases, while patterns associated with Type~B features may remain present in the robustified dataset, patterns associated with Type~C features, which were present in the original dataset, may be progressively removed, causing a detriment to accuracy as the robustness parameter increases significantly. Similarly, we speculate that this may also explain the well-known decrease in accuracy on original test data associated with an increase in model robustness as shown in Figure~\ref{fig:robust_test_accuracy}  \citep{tsipras2019robustness, zhang2019theoretically}.

Our results suggest that even when performance is critical and the drop in accuracy associated with adversarial training is unacceptable, adding even slight robustness ($\varepsilon=\nicefrac{1}{32}$) can drastically improve the universality of the learned features of a neural network. Networks can be improved in this way with little to no additional computational cost due to recent advances in improving the efficiency of adversarial training \citep{shafahi2019adversarial, wong2020fast, jeddi2020simple}.

We repeat these experiments using ResNet20 classifiers trained on the ImageNet-9 dataset and find similar results, presented in the Appendix.

\subsection{Transferability of Adversarial Examples} 

\begin{figure}[t]
\centering
\includegraphics[width=1.0\linewidth]{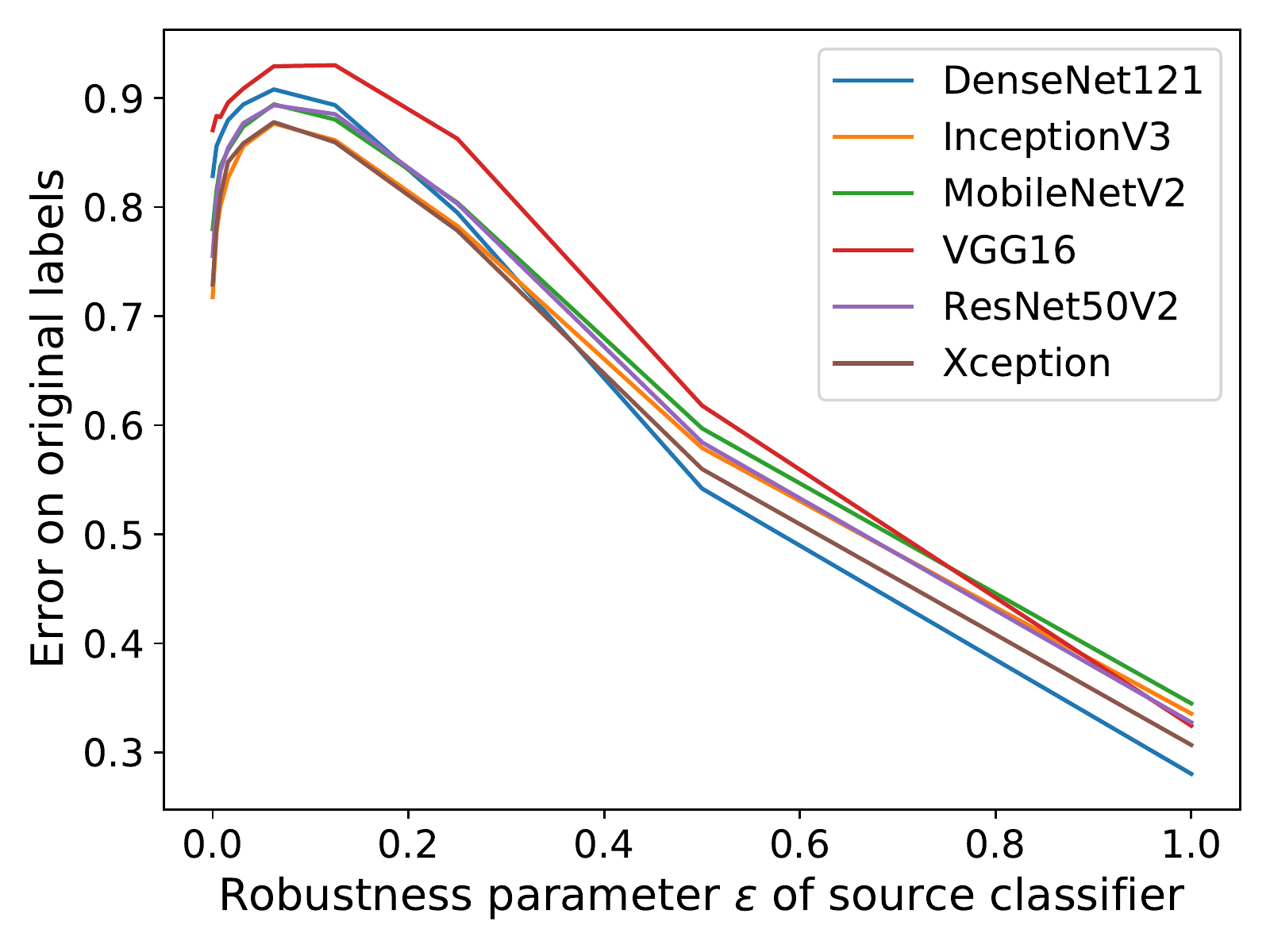}
\caption{Error rate of different standard classifiers evaluated on untargeted transfer attacks.  Each error rate is measured on a set of untargeted adversarial examples generated from the CIFAR-10 test set to attack a source classifier (ResNet50) with robustness parameter $\varepsilon$. Each adversarial example was generated with a perturbation $\delta$ where $\norm{\delta}_2 < 2.$ A higher error corresponds to a stronger transfer attack. See Appendix for the analogous graph for ImageNet-9. (Best viewed in color.)}
\label{subfig:untargeted_20}
\end{figure}

\begin{figure}[t]
\centering
\includegraphics[width=1.0\linewidth]{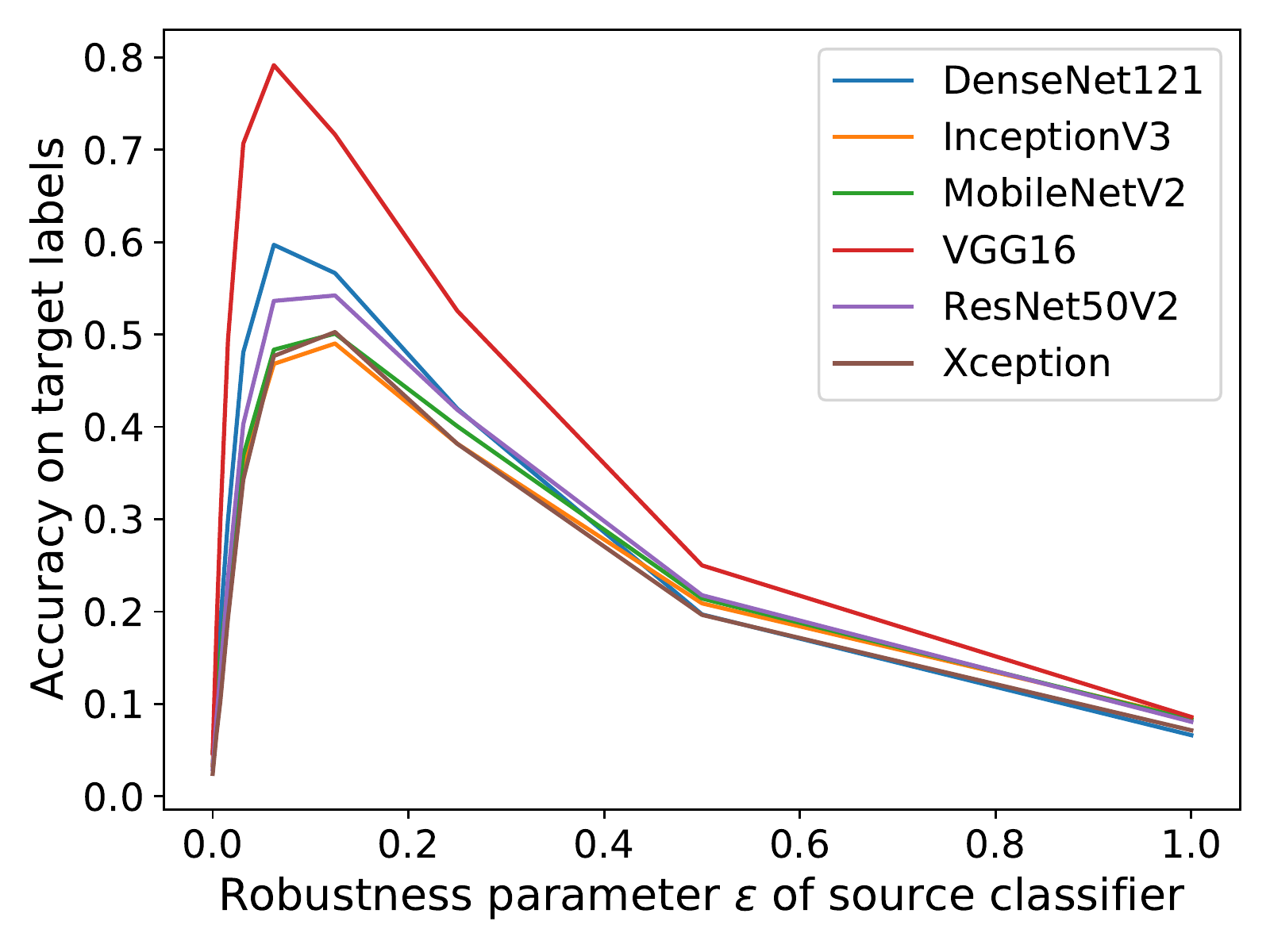}
\caption{Accuracy of different standard classifiers on targeted transfer attacks.  Each accuracy is measured as the fraction of times the target label $y_t$ is predicted for targeted adversarial example $x_t$.  The evaluation is done on a set of targeted adversarial examples generated from the CIFAR-10 test set to attack a source ResNet50 classifier with robustness parameter $\varepsilon$. Each adversarial example was generated with a perturbation $\delta$ where $\norm{\delta}_2 < 2.$ A higher accuracy corresponds to a stronger transfer attack. See Appendix for the analogous graph for ImageNet-9. (Best viewed in color.)}
\label{subfig:targeted_20}
\end{figure}

Adversarial examples designed to fool a given classifier are sometimes {\it transferable}: they can also successfully fool other classifiers, even those with different architectures from the original (``source'') classifier.  This is true for both \textit{untargeted} adversarial examples $x_{u}$, which are considered to be successful if a classifier predicts any incorrect label when given $x_{u}$, and for \textit{targeted} adversarial examples $x_{t}$, which are considered to be successful if a classifier predicts a specific targeted (incorrect) label $y_{t}$ when given $x_{t}$.  It is typically much harder to design transferable targeted than untargeted adversarial examples \citep{liu2016delving}.

Our earlier hypothesis---that robust features are more universal than non-robust features---suggests that adversarial examples designed to fool a robust classifier transfer (i.e., that exploit robust features) should be more transferable than adversarial examples designed to fool a non-robust classifier.  In this section we give evidence for this conclusion: we show that if a robust classifier is used as a source model for creating adversarial examples, those examples transfer more effectively than if a non-robust classifier is used as the source. This is an important novel result for researchers trying to create transferable adversarial examples: one should use a robust model as the source, instead of a standard model. 


To give evidence for these hypotheses, we train ResNet50 classifiers on the CIFAR-10 training set using adversarial training with varying robustness parameters $\varepsilon$ (in addition, we train ResNet20 on ImageNet-9 data, see Appendix).  We then use these robust classifiers to generate targeted and untargeted adversarial examples via projected gradient descent \citep{madry2017towards}, using the standard adversarial objectives. In the untargeted case, given an initial input $x$ and a parameter $\varepsilon$ to specify the maximum L2 norm of a permissible perturbation $\delta$, we seek to find an adversarial example $x_u = x + \hat{\delta}$ (where $x$ is the original example with label $y$) such that \[ \hat{\delta} = \argmax_{\norm{\delta}_2 \leq \varepsilon} \mathcal{L}(\theta, x + \delta, y), \] and for the targeted case, for a target label $y_t$, we seek to find an adversarial example $x_t = x + \hat{\delta}$ such that   \[ \hat{\delta} =  \argmin_{\norm{\delta}_2 \leq \varepsilon } \mathcal{L}(\theta, x + \delta, \hat{y}). \]
We evaluate the generated adversarial examples on standard classifiers with different architectures \citep{simonyan2014very, szegedy2016rethinking, howard2017mobilenets, huang2017densely, chollet2017xception}. For the untargeted case, Figure~\ref{subfig:untargeted_20} plots, for each classifier, the error rate on the original labels $y$, given adversarial examples $x_u$.  This error rate is plotted as a function of the robustness parameter $\varepsilon$ of the source classifier used to generate the adversarial examples ($x_u$).  The higher the error rate, the more successful the transfer attack.  To evaluate targeted adversarial examples, in Figure~\ref{subfig:targeted_20} we plot the accuracy of each standard classifier on adversarial examples $x_t$ with respect to the target label $y_t$---i.e., the fraction of times the classifier predicted the target label $y_t$, given example $x_t$.  The higher this accuracy, the more successful the transfer attack.

Figure~\ref{subfig:untargeted_20} shows that untargeted adversarial examples benefit slightly if generated with respect to a robust classifier with a small robustness parameter, $0 < \varepsilon \leq \nicefrac{1}{8}$ with a peak at $\varepsilon \approx \nicefrac{1}{16}$. The success of transferable adversarial examples decreases as $\varepsilon$ is further increased.

Figure~\ref{subfig:targeted_20} shows that transferable targeted adversarial examples attain a more striking increase in success by increasing the robustness parameter of the source classifier. Our technique for generating transferable targeted adversarial examples is largely ineffective when the source classifier is equivalent to standard classifier (i.e., $\varepsilon = 0$). However, using a robust source classifier with $\varepsilon \approx \nicefrac{1}{16}$ dramatically increases the success of targeted transfer attacks.

Experiments on the ImageNet-9 dataset using ResNet20 as source models behave similarly (see Appendix for details). 

Our goal here is not to generate state-of-the-art transfer attacks; we leave this to future work. Thus, we do not employ complex state-of-the-art transfer attacks and instead opt for the simple projected gradient descent approach. Nonetheless, our results confirm that robust features with small robustness parameter have high universality and can improve transfer attacks, especially in the targeted case, in which improvements are dramatic.

\section{Related Work}

Adversarial examples and associated adversarial robustness have been studied extensively in the machine learning literature \citep{goodfellow2014explaining, moosavi2016deepfool, allen2020feature, madry2017towards, engstrom2019adversarial, santurkar2019image, kaur2019perceptually, szegedy2014intriguing, biggio2013evasion, papernot2016limitations, warde201611, carlini2017towards, metzen2017detecting, feinman2017detecting, carlini2017adversarial, athalye2018synthesizing, carlini2019evaluating, cohen2019certified, uesato2018adversarial, stutz2019disentangling, shafahi2018adversarial, raghunathan2018certified}. Non-robust features have been studied directly by \citet{ilyas2019adversarial} who finds that non-robust features are present in image training datasets. By contrast, we establish an entanglement relationship between robust and non-robust features. 

We hypothesize that the overlap between robust and non-robust features may be explained in part from a simplicity bias \citep{valle2018deep, nakkiran2019sgd, shah2020pitfalls, arpit2017closer, wu2017towards, de2019random}, which suggests that neural networks learn simple functions more easily than complex functions, and from gradient starvation \citep{pezeshki2020gradient, combes2018learning}, which suggests that once highly predictive features are learned, other features become increasingly difficult to learn due to a diminishing gradient. If non-robust features are learned more readily and support sufficiently low classification loss, such a combination may impede the learning of robust features, for instance, see \citet{nakkiran2019adversarial}. Similarly, texture bias may be related \citep{hermann2020origins}.

Many papers have shown that deep learning classifiers may rely on non-semantic cues or shortcuts \citep{geirhos2020shortcut, mccoy2019right, jo2017measuring, wang2020high, wei2020understanding} and that optimization-based feature visualization may be inadequate to visualize neural network features \citep{borowski2020exemplary, hendrycks2019natural}. On the other hand, there has been work on feature visualization \citep{mahendran2015understanding, olah2017feature, zhang2018visual, dosovitskiy2016inverting, aubry2015understanding, simonyan2013deep}.

Related to our hypothesis that robust features are more universal than non-robust features, prior work has found that adversarial robustness serves as an effective prior for transfer learning \citep{terzi2020adversarial, salman2020adversarially, utrera2020adversarially, liang2020does}. There has been some work on the universality hypothesis \citep{li2015convergent, olah2020zoom, raghu2017svcca, kornblith2019similarity}. Additionally, there has been significant work on constructing and evaluating transferable targeted adversarial examples \citep{papernot2016transferability, moosavi2017universal, liu2016delving, kurakin2016adversarial, eykholt2018robust, tramer2017space, papernot2017practical}.

\section{Conclusion}

In this paper, we have presented an empirical study demonstrating that robust and non-robust features can be entangled in standard neural network classifiers. In particular, we have proposed three different types of features: robust features (Type A), non-robust features that are entangled with robust features (Type B), and non-robust features that are not entangled with robust features (Type C). Type A features are known to appear significantly more semantic than non-robust features \citep{engstrom2019adversarial,   santurkar2019image, kaur2019perceptually}. To our knowledge, no previous study has addressed the question of whether the non-robust features underlying adversarial vulnerability are the non-semantic yet highly predictive artifacts in image datasets, or if the features appear non-semantic for a different reason. We present evidence that at least some of the non-robust features indicate the presence of the same patterns of the dataset as robust features, suggesting that while appearing non-semantic when observed through the lens of their gradient and adversarial perturbations, these features may, in fact, indicate the presence of features that are aligned with human perception.

In addition, we provide evidence that robust features can be thought of as more universal than non-robust features, and as result, we find that robust classifiers are more effective as source classifiers for generating transferable adversarial examples in both the untargeted and targeted settings.

We believe that this work is an important step towards understanding the nature of non-robust features and answering the universality hypothesis \citep{olah2020zoom}. With a theory of robust and non-robust features, and thus of adversarial vulnerability and resilience, we hope to eventually understand how to build more interpretable classifiers and be able to better defend against adversarial attacks.

\section*{Acknowledgements}

Research presented in this article was supported by the Laboratory Directed Research and Development program of Los Alamos National Laboratory under project number 20210043DR.

The authors would like to thank Rory Soiffer for his helpful ideas and discussion.

\bibliography{references}
\bibliographystyle{icml2021}

\appendix

\onecolumn

\section{Experimental Setup}

In this section, we describe details of the methods that we use to train our standard, robust, zero-robust, and negative-robust models, construct robustified images, and construct adversarial examples.

\subsection{Dataset Considerations}

Our experiments involve training many models, a large portion of which require adversarial training. Similarly, our evaluation procedures involve constructing robust datasets, another highly computationally expensive operation. In order to ensure the computational feasibility of these problems, we conduct all experiments on two image datasets: CIFAR-10 \citep{krizhevsky2009learning} and a version of ImageNet \citep{ILSVRC15} which has been downsampled to $64 \times 64$ pixels \citep{chrabaszcz2017downsampled} and in which we only use images which fall under one of nine non-overlapping class labels: dog (classes 151--268), cat (classes 281--285), frog (classes 30--32), turtle (classes 33-37), bird (classes 80--100), primate (classes 365--382), fish (classes 389--397), crab (classes 118-121), insect (classes 300-319). This dataset is the same subset of the ImageNet dataset used by \citet{ilyas2019adversarial}, with the additional modification that our dataset is downsampled. CIFAR-10 has 50,000 training images and ImageNet-9 has approximately 200,000.

While we conduct our experiments only on these two datasets, we expect that our results should be general across all image datasets, and likely across other domains as well.

\subsection{Training \& Evaluation Details}

We describe the training procedure for the four different types of models that we construct: standard, robust, zero-robust, and negative-robust.

For each model, we train using the Adam optimizer \citep{kingma2014adam} with default TensorFlow parameters ($\beta_1 = 0.9, \beta_2=0.999$) \citep{tensorflow2015-whitepaper}. The learning rate for each model is specified in Table~\ref{tab:model_parameters}.  Each model was trained for at most 100 epochs. Models with an epoch count marked with an asterisks (*) had their training stopped early in order to maximize the performance on a validation set. Models that are not marked with an asterisks were trained for the entire 100 epochs.

\paragraph{Standard Models}
To train standard models, we minimize expected cross-entropy loss over the training dataset. All parameters are specified in Table~\ref{tab:model_parameters}. 

\paragraph{Robust Models}
To train robust models, we compute a set of adversarial examples for each batch during training. We use projected gradient descent to construct these adversarial examples \citep{madry2017towards}. For adversarial examples constructed from CIFAR-10 images, we use 16 steps with a step size of 0.5. For adversarial examples constructed from ImageNet images, we use 8 steps with a step size of 1.0. For adversarial training, we train to minimize the expected cross-entropy loss of the network evaluated on adversarial examples labeled as the true label. All other parameters are specified in Table~\ref{tab:model_parameters}.

\paragraph{Zero-Robust Models}
To train zero-robust models, we apply the methods proposed in Section 3.2. In particular, we construct adversarial examples (see Table~\ref{tab:dataset_parameters} for parameters) using projected gradient descent to minimize the targeted adversarial loss functions \[ \min_{\norm{\delta}_2 \leq \varepsilon} \mathcal{L}(\theta, x + \delta, \hat{y}) \] where $x$ is the original image, $\varepsilon$ is the maximum allowed L2 norm difference between the adversarial example and the original image, and $\hat{y}$ is the targeted class. For each image in CIFAR-10, we construct ten adversarial examples, one targeting each of the ten classes, labeled as the target class. For each image in the ImageNet-9 dataset, we construct a single adversarial example with a target selected uniformly at random from the nine possible classes, similarly labeled as the target class. We train zero-robust models for both CIFAR-10 and ImageNet-9 using standard training on the appropriate constructed dataset. The hyperparameters for training the models can be found in Table~\ref{tab:model_parameters}.

\paragraph{Negative-Robust Models}
We train negative-robust models in the same way as we train zero-robust models, except with a slightly different dataset. For each image with label $y$ in the CIFAR-10 dataset (numbered 0--9 to represent each class in CIFAR-10), we construct a new dataset with an adversarial example targeting the class $y + 1 \mod 10$, and labeled as this target class. We construct the ImageNet-9 version similarly, using $y + 1 \mod 9$ as the target label instead. The parameters for constructing the dataset can be found in Table~\ref{tab:dataset_parameters} and the hyperparameters for training the models in Table~\ref{tab:model_parameters}.

\paragraph{Victim Models}
To test the transferability of adversarial examples generated on robust models, we train standard models with different architectures: DenseNet121, InceptionV3, MobileNetV2, VGG16, ResNet50V2, and Xception \citep{simonyan2014very, szegedy2016rethinking, howard2017mobilenets, huang2017densely, chollet2017xception}. We initialize these models with pre-trained weights from training on the original ImageNet dataset \citep{chollet2015keras}. For the hyperparameters, see Table~\ref{tab:model_parameters}.

\begin{table*}[h]
\centering
\caption{Model parameters for each model trained in this paper. Parameters were selected by hand. In addition to what is listed below, the zero- and negative- robust models for both ImageNet-9 and CIFAR-10 were stopped prior to 100 epochs in order to maximize accuracy on a validation set. Each model was trained on a IBM Power9 architecture machine with two NVIDIA Tesla V100 GPUs. Each standard model took approximately 24 hours to train. Each robust models took approximately 48 hours.}
\label{tab:model_parameters}
\begin{tabular}{@{}llllllll@{}} \toprule
Model                  & Architecture  & \begin{tabular}[c]{@{}l@{}}Batch\\ size\end{tabular} & Epochs & \begin{tabular}[c]{@{}l@{}}Learning\\ rate\end{tabular} & \begin{tabular}[c]{@{}l@{}}LR \\ decay\end{tabular} & \begin{tabular}[c]{@{}l@{}}Data\\ augmentation\end{tabular} & Pretraining \\ \midrule
Standard (CIFAR)       & ResNet56 (v2) & 200   & 100     & 1e-3          & Yes      & Yes    & No           \\
Robust (CIFAR)         & ResNet56 (v2) & 200   & 100     & 1e-3          & Yes      & Yes    & No           \\
Zero-robust (CIFAR)    & ResNet56 (v2) & 200   & 100*     & 1e-4          & Yes      & No    & No            \\
Neg-robust (CIFAR)     & ResNet56 (v2) & 200   & 100*     & 1e-4          & Yes      & No     & No           \\
Victim models (CIFAR) & Misc & 200 & 100 & 1e-4 & Yes & Yes & Yes \\ 
Standard (ImageNet)    & ResNet20 (v2) & 128   & 100     & 1e-3          & Yes      & Yes    & No           \\
Robust (ImageNet)      & ResNet20 (v2) & 128   & 100     & 1e-3          & Yes      & Yes    & No           \\
Zero-Robust (ImageNet) & ResNet20 (v2) & 128   & 100*     & 1e-3          & Yes      & No    & No            \\
Neg-robust (ImageNet)  & ResNet20 (v2) & 128   & 100*     & 1e-3          & Yes      & No     & No          \\
Victim models (ImageNet) & Misc & 128 & 100 & 1e-4 & Yes & Yes & Yes \\ \bottomrule
\end{tabular}
\end{table*}

\begin{table*}[h]
\centering
\caption{Parameters for the various datasets constructed in this paper. Parameters were selected by hand.}
\label{tab:dataset_parameters}
\begin{tabular}{@{}llll@{}} \toprule
Dataset                                                                               & \begin{tabular}[c]{@{}l@{}}Maximum allowed change\\ (L2 norm from seed)\end{tabular} & Number of steps & Step size \\ \midrule
Robustified (CIFAR)                                                                   & $\infty$                                                                             & 400             & 0.05      \\
Zero-robust (CIFAR)                                                                   & 0.5                                                                                  & 200             & 0.1       \\
Neg-robust (CIFAR)                                                                    & 0.5                                                                                  & 200             & 0.1       \\
\begin{tabular}[c]{@{}l@{}}Untargeted Adversarial Examples (CIFAR)\end{tabular}    & 2.0                                                                                  & 100             & 0.1       \\
\begin{tabular}[c]{@{}l@{}}Targeted Adversarial Examples (CIFAR)\end{tabular}      & 2.0                                                                                  & 100             & 0.1       \\
Robustified (ImageNet)                                                                & $\infty$                                                                             & 200             & 0.1       \\
Zero-robust (ImageNet)                                                                & 0.5                                                                                  & 200             & 0.1       \\
Neg-robust (ImageNet)                                                                 & 0.5                                                                                  & 200             & 0.1       \\
\begin{tabular}[c]{@{}l@{}}Untargeted Adversarial  Examples (ImageNet)\end{tabular} & 4.0                                                                                  & 100             & 0.1       \\
\begin{tabular}[c]{@{}l@{}}Targeted Adversarial Examples (ImageNet)\end{tabular}    & 4.0                                                                                  & 100             & 0.1      \\ \bottomrule
\end{tabular}
\end{table*}

\section{Extended Results}

\begin{figure*}[h]
\centering
\includegraphics[width=0.5\textwidth]{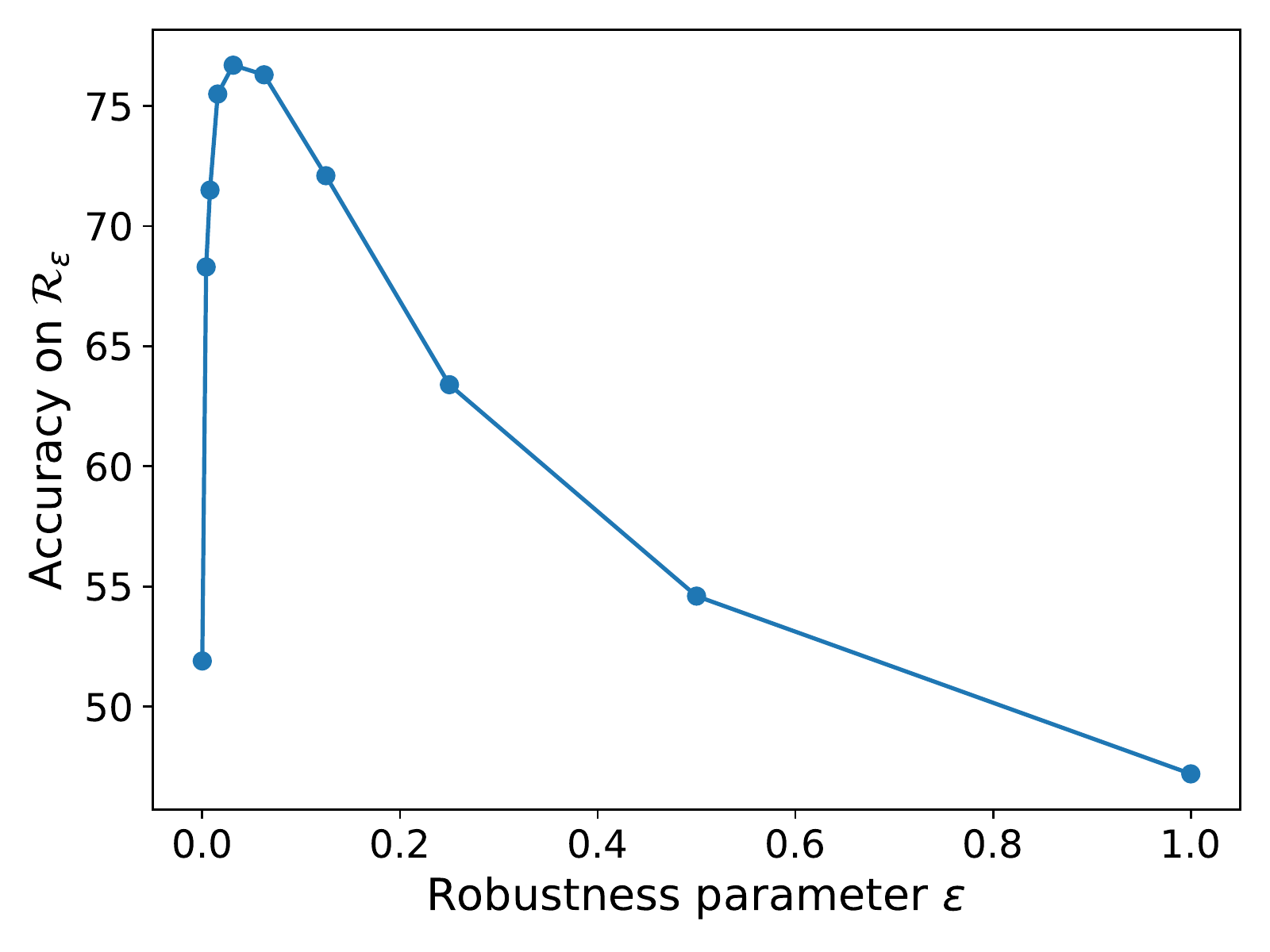}
\caption{Accuracy of a standard ResNet20 classifier on robustifications of the ImageNet-9 test dataset, generated with respect to robust ResNet20 classifiers with varying robustness parameter $\varepsilon$. (Best viewed in color.)}
\label{fig:standard_performance_on_robustified_imagenet}
\end{figure*}

\begin{figure}[h]
\includegraphics[width=\linewidth]{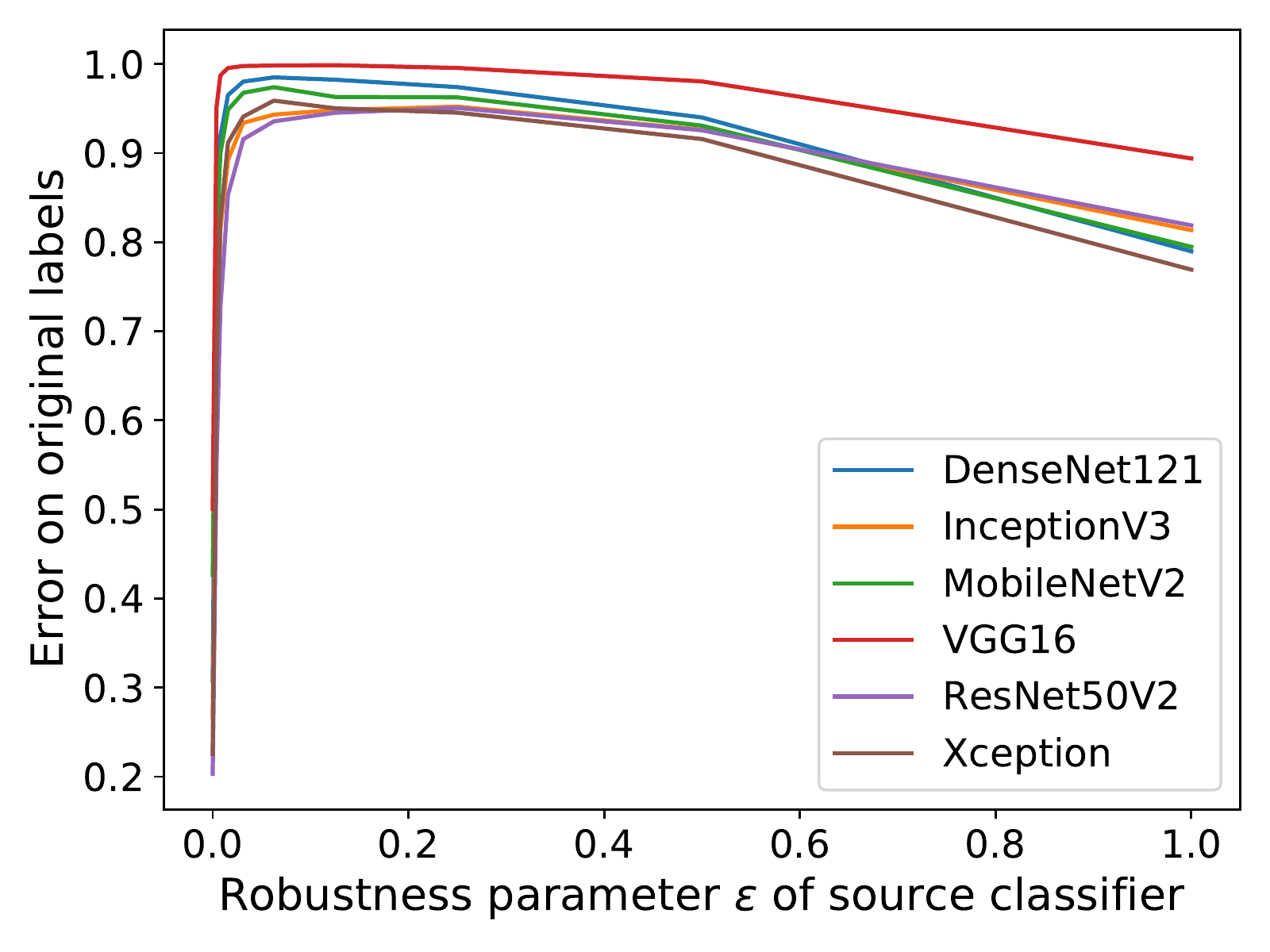}
\caption{Error rate of different standard classifiers evaluated on untargeted transfer attacks.  Each error rate is measured on a set of untargeted adversarial examples generated from the ImageNet-9 test set to attack a source classifier (ResNet20) with robustness parameter $\varepsilon$. Each adversarial example was generated with a perturbation $\delta$ where $\norm{\delta}_2 < 2.$ A higher error corresponds to a stronger transfer attack. (Best viewed in color.)}
\label{subfig:untargeted_imagenet}
\end{figure}

\begin{figure}[h]
\includegraphics[width=\linewidth]{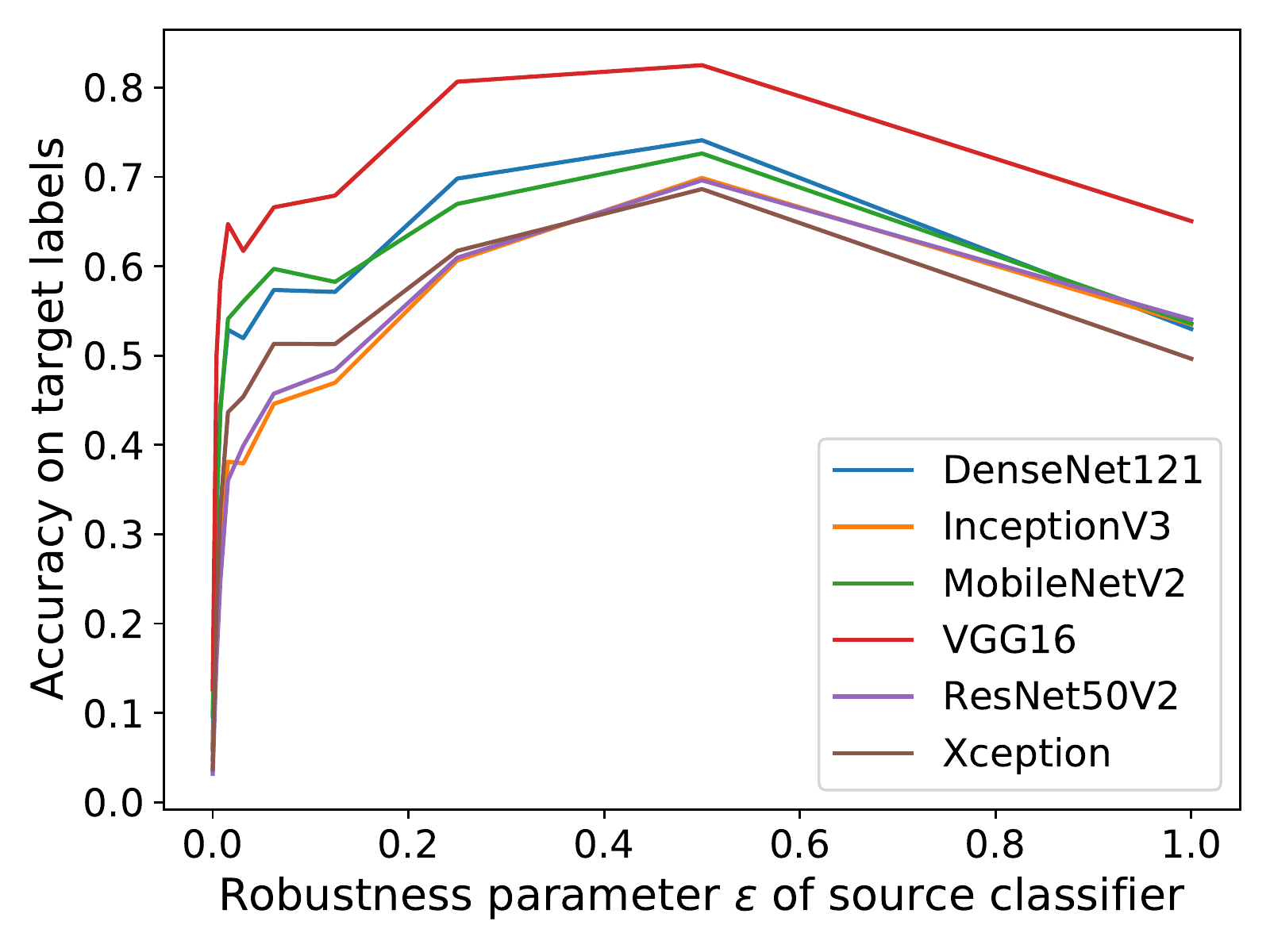}
\caption{Accuracy of different standard classifiers on targeted transfer attacks.  Each accuracy is measured as the fraction of times the target label $y_t$ is predicted for targeted adversarial example $x_t$.  The evaluation is done on a set of targeted adversarial examples generated from the ImageNet-9 test set to attack a source ResNet20 classifier with robustness parameter $\varepsilon$. Each adversarial example was generated with a perturbation $\delta$ where $\norm{\delta}_2 < 2.$ A higher accuracy corresponds to a stronger transfer attack. (Best viewed in color.)}
\label{subfig:targeted_imagenet}
\end{figure}

\begin{figure}[h]
    \centering
    \includegraphics[width=\linewidth]{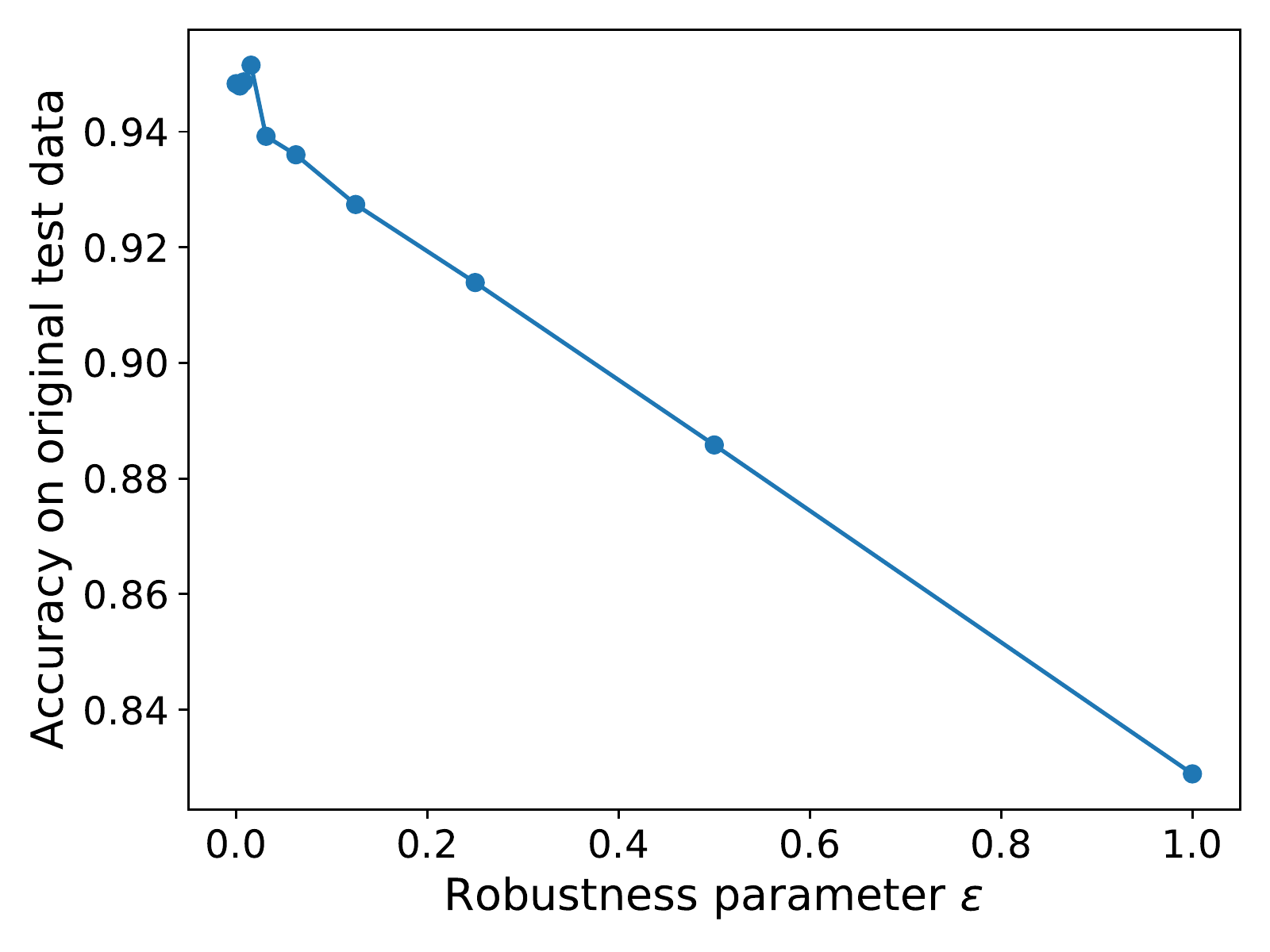}
    \caption{Accuracy of robust ResNet20s of varying robustness parameter trained on the ImageNet-9 training data and evaluated on the ImageNet-9 original test data. Confirms well known result: test accuracy decreases as robustness increases.}
    \label{fig:imagenet_robust_test_accuracy}
\end{figure}

\begin{figure*}[h]
    \centering
    \includegraphics[width=0.9\textwidth, trim = 0cm 2.5cm 0cm 4cm, clip]{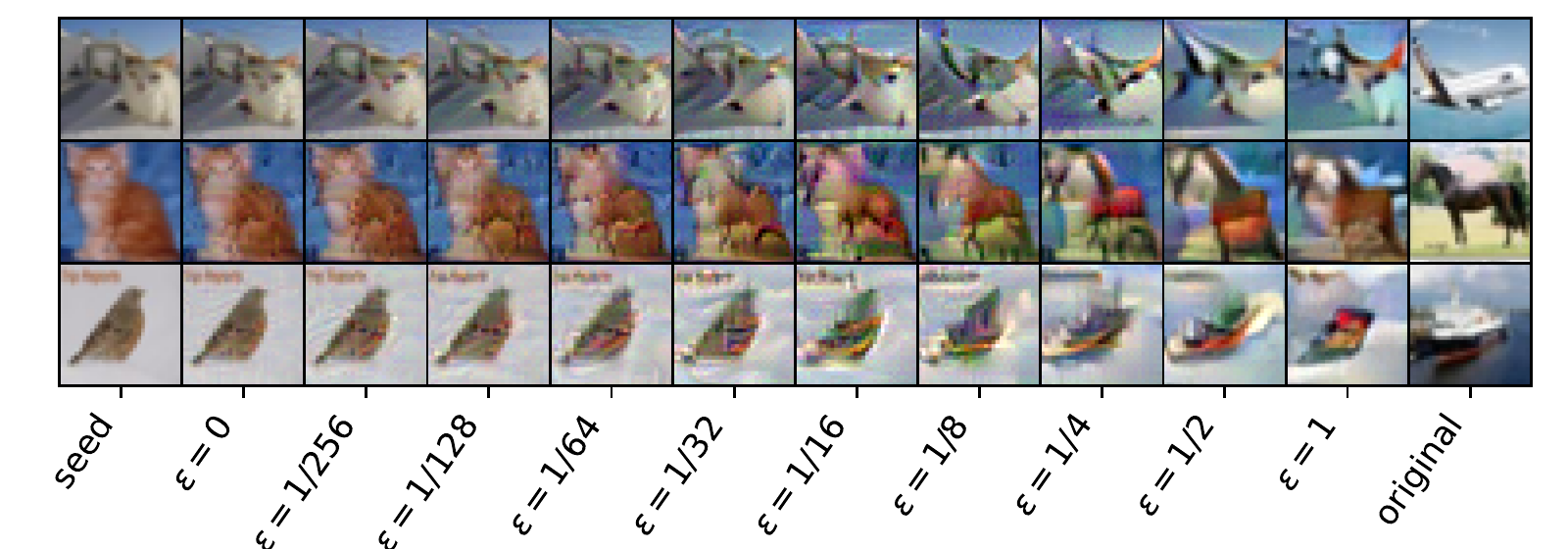}
    \caption{Extended data: Examples of robustified images by $\varepsilon$ of robust classifier. The rightmost column depicts the original CIFAR-10 image; the leftmost column depicts the seed ($x_0$) from which the gradient descent of the robustification process began; the rest are robustified images. The representation layer of an $\varepsilon$-robust classifier for each specified $\varepsilon$ is approximately the same for the robustified image and the associated original image. Since the representation layers of less robust classifiers are more easily perturbed, robustified images generated with respect to classifiers with a smaller robustness parameter appear closer to the (random) CIFAR-10 image that seeded the gradient-descent process. (Best viewed in color.)}
    \label{fig:robustification_examples_long}
\end{figure*}

\begin{figure*}[h]
    \centering
    \includegraphics[width=0.9\textwidth, trim = 0cm 2.5cm 0cm 4cm, clip]{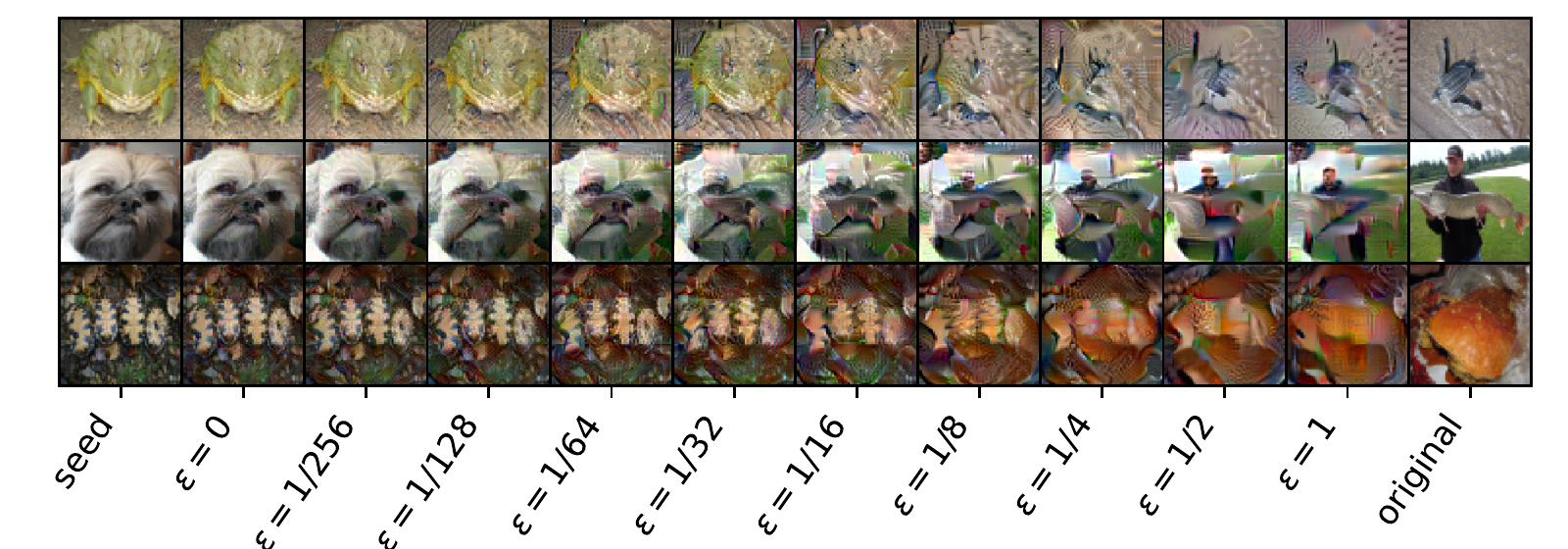}
    \caption{Extended data: Examples of robustified images by $\varepsilon$ of robust classifier. The rightmost column depicts the original ImageNet-9 image; the leftmost column depicts the seed ($x_0$) from which the gradient descent of the robustification process began; the rest are robustified images. The representation layer of an $\varepsilon$-robust classifier for each specified $\varepsilon$ is approximately the same for the robustified image and the associated original image. Since the representation layers of less robust classifiers are more easily perturbed, robustified images generated with respect to classifiers with a smaller robustness parameter appear closer to the (random) ImageNet-9 image that seeded the gradient-descent process. (Best viewed in color.)}
    \label{fig:imagenet_robustification_examples}
\end{figure*}

\begin{figure*}[h]
    \centering
    \includegraphics[width=0.9\textwidth, trim = 0cm 0cm 0cm 0cm, clip]{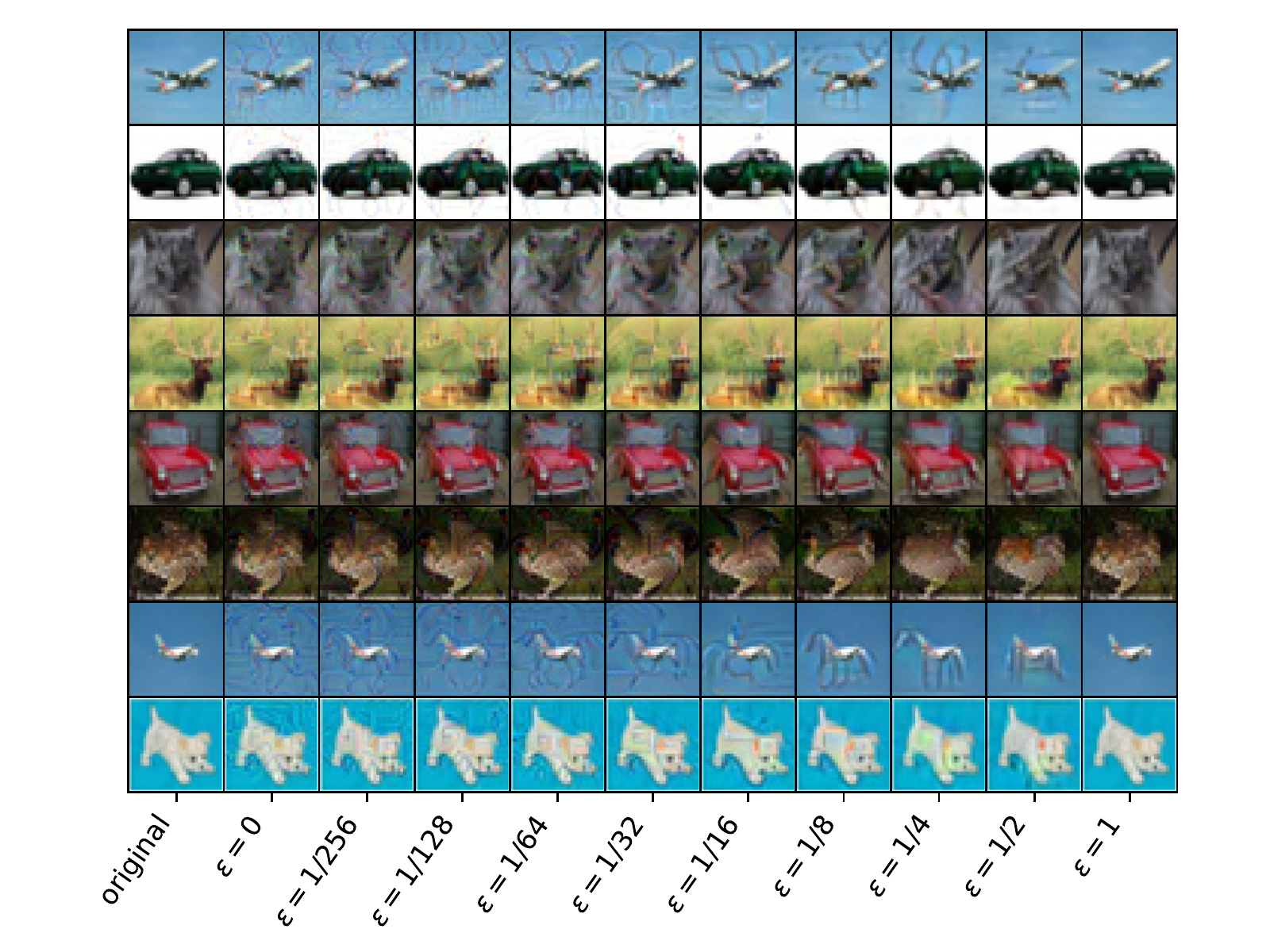}
    \caption{Extended data: Examples of adversarial examples for robust CIFAR-10 classifiers, of varying robustness. The left-most image is the original image from CIFAR-10. All other images are adversarial. Each adversarial example was generated via PGD from the original image $x$ such that the resulting adversarial example $\hat{x} = x + \delta$ where $\norm{\delta}_2 \leq 2$ (Best viewed in color.)}
    \label{fig:cifar_adversarial}
\end{figure*}

\begin{figure*}[h]
    \centering
    \includegraphics[width=0.9\textwidth, trim = 0cm 0cm 0cm 0cm, clip]{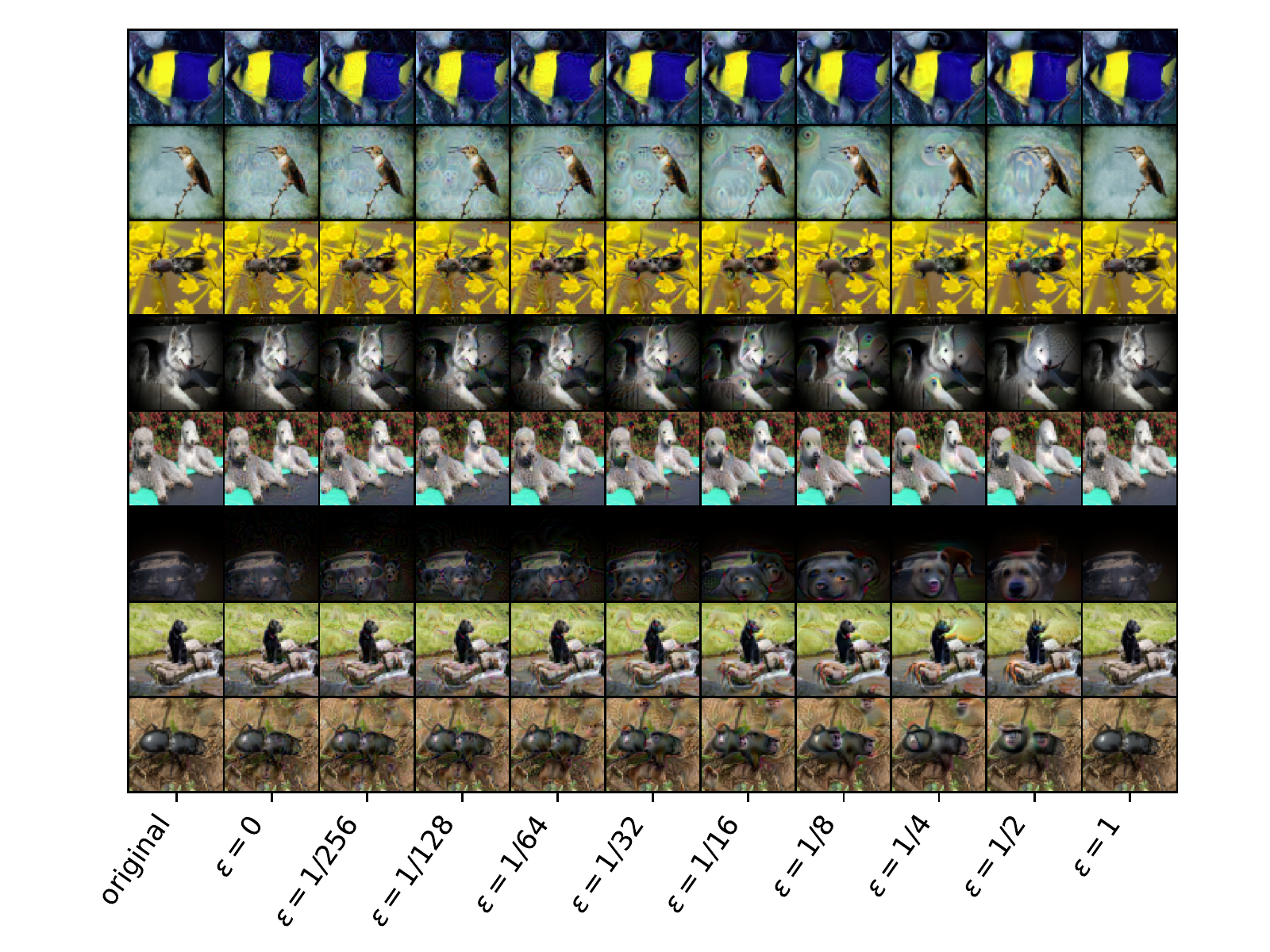}
    \caption{Extended data: Examples of adversarial examples for robust ImageNet-9 classifiers, of varying robustness. The left-most image is the original image from ImageNet-9. All other images are adversarial. Each adversarial example was generated via PGD from the original image $x$ such that the resulting adversarial example $\hat{x} = x + \delta$ where $\norm{\delta}_2 \leq 4$ (Best viewed in color.)}
    \label{fig:imagenet_adversarial}
\end{figure*}

In this section, we present extended results including the extended data from Tables~\ref{tab:performance_robustified}~and~\ref{tab:performance_robustified_imagenet} (Tables~\ref{tab:ex_performance_robustified}~and~\ref{tab:ex_performance_robustified_imagenet}). In addition, we include ImageNet-9 versions of Figures~\ref{fig:standard_performance_on_robustified},~\ref{subfig:untargeted_20},~and~\ref{subfig:targeted_20} (Figures~\ref{fig:standard_performance_on_robustified_imagenet},~\ref{subfig:targeted_imagenet},~and~\ref{subfig:untargeted_imagenet}). We present the accuracy of each of our robust classifiers on the original ImageNet-9 testing set (Figure~\ref{fig:imagenet_robust_test_accuracy}). We include extended examples of robustified images (Figures~\ref{fig:robustification_examples_long}~and~\ref{fig:imagenet_robustification_examples}). Finally, we provide examples of adversarial examples generated on robust models (Figures~\ref{fig:cifar_adversarial}~and~\ref{fig:imagenet_adversarial}).

\begin{table*}[h]
\centering
\caption{Extended data: Accuracy of zero-robust, negative-robust, standard, and robust ResNet50 classifiers on robustified CIFAR-10 test set. All classifiers are trained starting with random initial weights. The \textit{Test} column is the accuracy on the original CIFAR-10 test set. The $\mathcal{R}_\varepsilon$ columns give accuracy on robustified CIFAR-10 test set, generated with respect to a robust ResNet50 with robustness parameter $\varepsilon$.}
\label{tab:ex_performance_robustified}
\begin{tabular}{@{}llllllllllll@{}}
\toprule
Classifier & Test & $R_0$ & $\mathcal{R}_{\nicefrac{1}{256}}$ & $\mathcal{R}_{\nicefrac{1}{128}}$ & $\mathcal{R}_{\nicefrac{1}{64}}$ & $\mathcal{R}_{\nicefrac{1}{32}}$ & $\mathcal{R}_{\nicefrac{1}{16}}$ & $\mathcal{R}_{\nicefrac{1}{8}}$ & $\mathcal{R}_{\nicefrac{1}{4}}$ & $\mathcal{R}_{\nicefrac{1}{2}}$ & $\mathcal{R}_{1}$ \\ \midrule
Standard & 94.1 & 58.3 & 77.3 & 81.4 & 82.2 & 83.7 & 79.9 & 79.0 & 75.1 & 69.2 & 56.3 \\
Robust ($\varepsilon=0.5$) & 86.5 & 9.93 & 9.97 & 9.96 & 10.0 & 11.1 & 18.2 & 44.5 & 70.5 & 86.5 & 70.0 \\
Zero-robust & 46.8 & 38.8 & 35.1 & 26.1 & 27.2 & 26.2 & 23.3 & 22.6 & 22.1 & 23.4 & 21.4 \\
Negative-robust & 21.9 & 31.8 & 15.8 & 14.6 & 16.6 & 11.3 & 11.1 & 11.0 & 12.8 & 14.3 & 13.2 \\
\bottomrule
\end{tabular}
\end{table*}

\begin{table*}[h]
\centering
\caption{Extended data: Accuracy of zero-robust, negative-robust, standard, and robust ResNet20 classifiers on robustified ImageNet-9 test set. Analogous to Table~\ref{tab:ex_performance_robustified}. Note: since there are nine classes, chance accuracy is approximately $11.1\%$.}
\label{tab:ex_performance_robustified_imagenet}
\begin{tabular}{@{}llllllllllll@{}}
\toprule
Classifier & Test & $R_0$ & $\mathcal{R}_{\nicefrac{1}{256}}$ & $\mathcal{R}_{\nicefrac{1}{128}}$ & $\mathcal{R}_{\nicefrac{1}{64}}$ & $\mathcal{R}_{\nicefrac{1}{32}}$ & $\mathcal{R}_{\nicefrac{1}{16}}$ & $\mathcal{R}_{\nicefrac{1}{8}}$ & $\mathcal{R}_{\nicefrac{1}{4}}$ & $\mathcal{R}_{\nicefrac{1}{2}}$ & $\mathcal{R}_{1}$ \\ \midrule
Standard & 94.8 & 51.9 & 68.3 & 71.5 & 75.5 & 76.7 & 76.3 & 72.1 & 63.4 & 54.6 & 47.2 \\
Robust ($\varepsilon=1.0$) & 82.9 & 41.9 & 42.1 & 42.5 & 42.5 & 42.5 & 44.1 & 46.1 & 49.0 & 51.7 & 56.0 \\
Zero-robust & 49.2 & 24.8 & 28.3 & 30.5 & 29.3 & 27.5 & 24.3 & 24.2 & 24.6 & 28.6 & 31.4 \\
Negative-robust  & 38.3 & 25.4 & 18.8 & 18.4 & 19.2 & 20.2 & 23.2 &  23.6 & 21.8 & 24.4 & 23.7 \\
\bottomrule
\end{tabular}
\end{table*}

\end{document}